\documentclass[10pt,twocolumn,letterpaper]{article}

\usepackage[pagenumbers]{cvpr} 

\usepackage{graphicx}
\usepackage{amsmath}
\usepackage{amssymb}
\usepackage{booktabs}
\usepackage[normalem]{ulem}
\usepackage[accsupp]{axessibility}

\usepackage{enumitem}

\usepackage[pagebackref,breaklinks,colorlinks]{hyperref}

\usepackage[capitalize]{cleveref}
\crefname{section}{Sec.}{Secs.}
\Crefname{section}{Section}{Sections}
\Crefname{table}{Table}{Tables}
\crefname{table}{Tab.}{Tabs.}

\def\confName{CVPR}
\def\confYear{2023}

\def\ModelName{FitMe}
\def\GANName{BRDF-GAN}

\begin{document}

\title{\ModelName{}: Deep Photorealistic 3D Morphable Model Avatars}

\author{Alexandros Lattas
       \hspace{0.8cm}
       Stylianos Moschoglou
       \hspace{0.8cm}
       Stylianos Ploumpis
       \\
       Baris Gecer
       \hspace{0.8cm}
       Jiankang Deng
       \hspace{0.8cm}
       Stefanos Zafeiriou
\and
Imperial College London, UK
\\
{\tt\small\{a.lattas,s.moschoglou,s.ploumpis,b.gecer,j.deng16,s.zafeiriou\}@imperial.ac.uk}
}

\twocolumn[{%
\renewcommand\twocolumn[1][h]{#1}%
\maketitle
\begin{center}
    \centering
    \vspace{-0.35cm}
    \captionsetup{type=figure}
    \includegraphics[width=\textwidth]{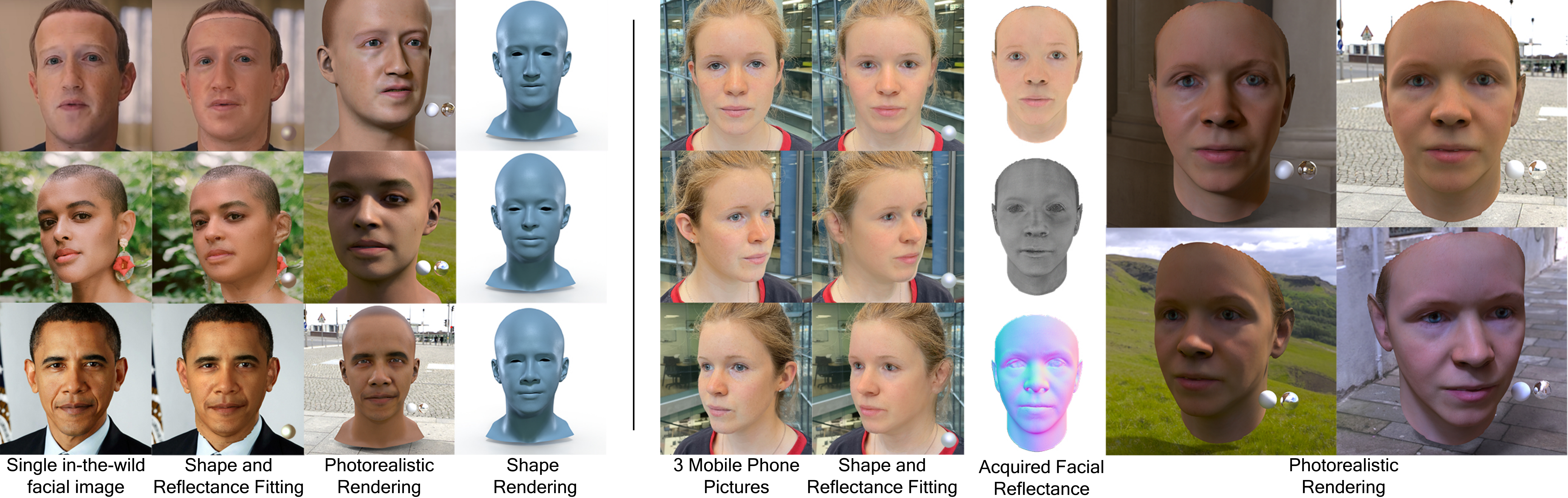}\\
    \captionof{figure}{
        \ModelName{} reconstructs relightable shape and reflectance maps for facial avatars,
        from a single (left) or multiple (right) unconstrained facial images,
        using a reflectance model and differentiable rendering.
        The results can be photorealistically rendered in common engines.
    }
    \label{fig:teaser}
\end{center}%
}]

\begin{abstract}
  In this paper, 
  we introduce \textit{\ModelName{}},
  a facial reflectance model and a differentiable rendering optimization pipeline,
  that can be used to acquire high-fidelity renderable human avatars from single or multiple images.
  The model consists of a multi-modal style-based generator,
  that captures facial appearance in terms of diffuse and specular reflectance,
  and a PCA-based shape model.
  We employ a fast differentiable rendering process that can be used in an optimization pipeline,
  while also achieving photorealistic facial shading.
  Our optimization process accurately captures both the facial reflectance and shape in high-detail,
  by exploiting the expressivity of the style-based latent representation and of our shape model.
  \ModelName{} achieves state-of-the-art reflectance acquisition and identity preservation
  on single ``in-the-wild'' facial images,
  while it produces impressive scan-like results, 
  when given multiple unconstrained facial images pertaining to the same identity.
  In contrast with recent implicit avatar reconstructions,
  \ModelName{} requires only one minute 
  and produces relightable mesh and texture-based avatars, 
  that can be used by end-user applications.
  Project page at \url{lattas.github.io/fitme}.
\end{abstract}

\section{Introduction}
\label{sec:introduction}

\begin{figure*}[ht]
  \centering
  \includegraphics[width=\linewidth]{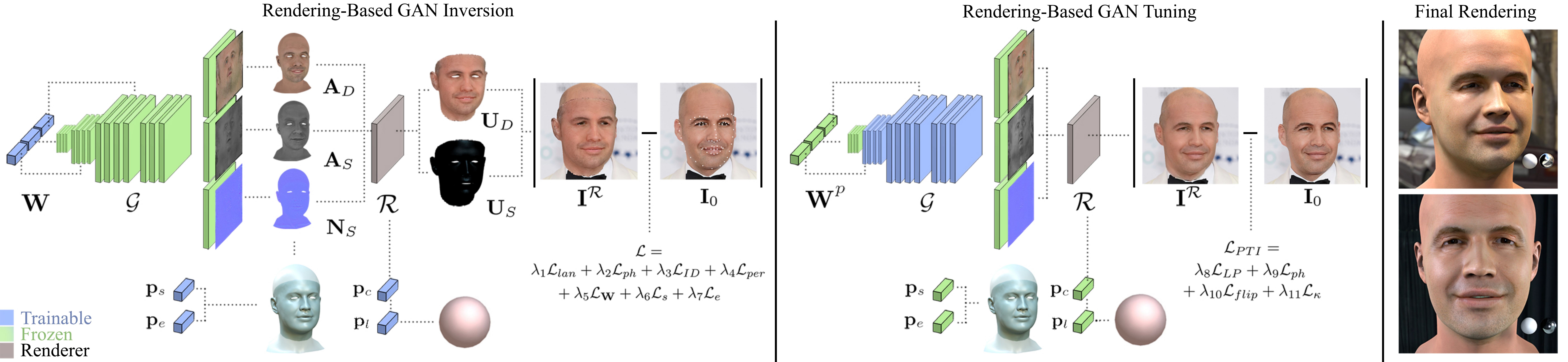}
  \caption{
    \ModelName{} method overview.
    For a target image $\mathbf{I}_0$, we optimize the latent vector $\mathbf{W}$ of the generator $\mathcal{G}$,
    and the shape identity $\mathbf{p}_s$, expression $\mathbf{p}_e$, camera $\mathbf{p}_c$ and illumination $\mathbf{p}_l$ parameters, by combining 3DMM fitting and GAN inversion methods,
    through accurate differentiable diffuse $\mathbf{U}_D$ and specular $\mathbf{U}_S$ rendering $\mathcal{R}$.
    Then, from the optimized $\mathbf{W}^p$, we tune the generator $\mathcal{G}$ weights,
    through the same rendering process.
    The reconstructed shape $\mathbf{S}$ and facial reflectance (diffuse albedo $\mathbf{A}_D$,
    specular albedo $\mathbf{A}_S$ and normals $\mathbf{N}_S$), achieve great identity similarity
    and can be directly used in typical renderers,
    to achieve photorealistic rendering, as shown on the right-hand side.
  }
  \label{fig:method_overview}
\end{figure*}

Despite the tremendous steps forward witnessed in the last decade, 3D facial reconstruction from a single unconstrained image remains an important research problem 
with an active presence in the computer vision community.
Its applications are now wide-ranging, including but not limited to human digitization 
for virtual and augmented reality applications,
social media and gaming, synthetic dataset creation, and health applications.
However, recent works come short of accurately reconstructing 
the identity of different subjects and usually fail to produce
assets that can be used for photorealistic rendering.
This can be attributed to the lack of diverse and big datasets 
of scanned human geometry and reflectance,
the limited and ambiguous information available on a single facial image,
and the limitations of the current statistical and machine learning methods.

3D Morhpable Models (3DMM) \cite{egger20203d} have been a standard method of facial shape and appearance acquisition
from a single ``in-the-wild'' image.
The seminal 3DMM work in \cite{blanz1999morphable} used Principal Component Analysis (PCA),
to model facial shape and appearance with variable identity and expression,
learned from about 200 subjects.
Since then, larger models have been introduced,
.i.e.~the LSFM \cite{booth20163d}, Basel Face Model \cite{paysan20093d}
and Facescape \cite{yang2020facescape}, with thousands of subjects.
Moreover, recent works have introduced 3DMMs of complete human heads 
\cite{li2017learning, ploumpis2019combining, ploumpis2020towards} 
or other facial parts such as ears \cite{ploumpis2020towards} and tongue \cite{ploumpis20223d}. 
Finally, recent works have introduced extensions ranging from non-linear models
\cite{moschoglou20203dfacegan, tran2018nonlinear, tran2019towards} 
to directly regressing 3DMM parameters \cite{tuan2017regressing, sanyal2019learning}.
However, such models cannot produce textures capable 
of photorealistic rendering.

During the last decade we have seen considerable improvements in deep generative models.
Generative Adversarial Networks (GANs) \cite{goodfellow_generative_2014},
and specifically progressive GAN architectures \cite{karras2018progressive}
have achieved tremendous results in learning distributions of high-resolution 2D images of human faces.
Recently, style-based progressive generative networks \cite{karras2019style,karras2020training,karras2020analyzing,karras2021alias}
are able to learn meaningful latent spaces, that can be traversed in order to reconstruct
and manipulate different attributes of the generated samples.
Some methods have also been shown effective in learning a 2D representation of 3D facial attributes,
such as UV maps \cite{gecer2019ganfit, gecer2021fast, luo2021normalized, gecer2021ostec}.

3D facial meshes generated by 3DMMs can be utilized in rendering functions,
in order to create 2D facial images.
Differentiating the rendering process is also required in order to perform iterative optimization.
Recent advances in differentiable rasterization \cite{liu2019soft},
photorealitic facial shading \cite{lattas2021avatarme++} and rendering libraries \cite{ravi2020pytorch3d, Genova_2018_CVPR, KaolinLibrary},
enable the photorealistic differentiable rendering of such assets.
Unfortunately, 3DMM \cite{booth20163d, gecer2019ganfit, luo2021normalized} works rely on the lambertian shading model
which comes short of capturing the complexity of facial reflectance.
The issue being, photorealistic facial rendering requires various facial reflectance parameters
instead of a single RGB texture \cite{lattas2021avatarme++}.
Such datasets are scarce, small and difficult to capture \cite{ma2007rapid,ghosh2011multiview,riviere2020single},
despite recent attempts to simplify such setups \cite{lattas2022practical}.

Several recent approaches have achieved either high-fidelity facial reconstructions
\cite{gecer2019ganfit, luo2021normalized, bao2021high}
or relightable facial reflectance reconstructions \cite{smith2020morphable, dib2021towards, dib2021practical, lattas2020avatarme, lattas2021avatarme++, feng2022towards}, including infra-red \cite{miao2022physicallybased},
however, high-fidelity and relightable reconstruction still remains elusive.
Moreover, powerful models have been shown to capture facial appearance with deep models
\cite{li2020learning, gecer2020synthesizing}, but they fail to show single or multi image reconstructions.
A recent alternative paradigm uses implicit representations
to capture avatar appearance and geometry \cite{cao2022authentic, wang2022morf},
the rendering of which depends on a learned neural rendering.
Despite their impressive results, such implicit representations
cannot be used by common renderers and are not usually relightable.
Finally, the recently introduced Albedo Morphable Model (AlbedoMM) \cite{smith2020morphable} 
captures facial reflectance and shape with a linear PCA model,
but per-vertex color and normal reconstruction is too low-resolution for photorealistic rendering.
AvatarMe++ \cite{lattas2020avatarme, lattas2021avatarme++}
reconstructs high-resolution facial reflectance texture maps from a single ``in-the-wild'' image,
however, its 3-step process (reconstruction, upsampling, reflectance),
cannot be optimized directly with the input image.

In this work, we introduce \textit{\ModelName{}},
a fully renderable 3DMM with high-resolution facial reflectance texture maps,
which can be fit on unconstrained facial images using accurate differentiable renderings.
\ModelName{} achieves identity similarity and high-detailed, fully renderable reconstructions,
which are directly usable by off-the-shelf rendering applications.
The texture model is designed as a multi-modal style-based progressive generator, which concurrently generates the facial \textit{diffuse-albedo}, \textit{specular-albedo}
and \textit{surface-normals}.
A meticulously designed branched discriminator enables smooth training with modalities of different statistics.
To train the model we create a capture-quality facial reflectance dataset of 5k subjects,
by fine-tuning AvatarMe++ on the MimicMe \cite{papaioannou2022mimicme} public dataset, which we also augment in order to balance skin-tone representation.
For the shape, we use interchangeably a face
and head PCA model \cite{ploumpis2020towards},
both trained on large-scale geometry datasets.
We design a single or multi-image fitting method,
based on style-based generator projection \cite{karras2020training} and 3DMM fitting.
To perform efficient iterative fitting (in under 1 minute), 
the rendering function needs to be differentiable and fast, 
which makes models such as path tracing unusable. 
Prior works in the field \cite{booth20163d,chen2019photo,gecer2019ganfit}
use simpler shading models (e.g. Lambertian), or much slower optimization \cite{dib2021practical}.
We add a more photorealistic shading than prior work,
with plausible diffuse and specular rendering,
which can acquire shape and reflectance capable of photorealistic rendering
in standard rendering engines (Fig.~\ref{fig:teaser}).
The flexibility of the generator's extended latent space and the photorealistic fitting,
enables \ModelName{} to reconstruct high-fidelity facial reflectance and achieve impressive identity similarity,
while accurately capturing details in diffuse, specular albedo and normals.
Overall, in this work we present:
\begin{itemize}
  \item the first 3DMM capable of generating high-resolution facial 
 reflectance and shape,
        with an increasing level of detail, that can be photorealistically rendered,
  \item the first branched multi-modal style-based progressive generator of high-resolution 3D facial assets (diffuse albedo, specular albedo and normals), and a suitable multi-modal branched discriminator,
  \item a method to acquire and augment a vast facial reflectance dataset of, using assets from a public dataset,
  \item a multi-modal generator projection, optimized with diffuse and specular differentiable rendering.
\end{itemize}


\section{Related Work}
\label{sec:related}

\subsection{3D Morphable Models}
\label{sec:related_3dmm}
Early facial modeling and fitting methods, starting from the seminal 3DMM work of Blanz and Vetter in \cite{blanz1999morphable}, have always used a linear model for facial shape and appearance \cite{paysan20093d,booth20163d}. They are thoroughly analyzed in the recent review of Egger et al.~\cite{egger20203d}. Moreover, 3DMMs have also been extended to the entire head in \cite{ploumpis2020towards}. 
Finally, the recent AlbedoMM \cite{smith2020morphable} uses separate PCA models of the diffuse and specular albedo, however the per-vertex albedo is low resolution and inadequate for photorealistic rendering.

Recent works have been replacing parts of the linear shape/texture models with neural networks in order to capture non-linearities. 
Bagautdinov et al.\cite{bagautdinov2018modeling} introduced the first method for non-linear facial shape modeling based on variational autoencoders. 3DFaceGAN \cite{moschoglou20203dfacegan} introduced a GAN-based approach for facial geometry based on UV maps which captured important non-linearities of the facial shape. Closest to our work, \cite{gecer2020synthesizing, li2020learning} designed models that combines facial albedo, shape, and normals. 

\subsection{Deep Generative Networks}
\label{sec:related_stylegan}
GANs \cite{goodfellow_generative_2014} first achieved high-resolution facial generation
with the progressive growing of GANs \cite{karras2018progressive}, 
which introduced a generator-discriminator pair trained on progressively growing resolutions.
StyleGAN \cite{karras2019style} also introduced a noise-injection technique and a mapping network that learns meaningful latent representations. 
StyleGAN2 \cite{karras2020analyzing} further optimized the architecture and projection method,
while StyleGAN2ADA \cite{karras2020training} 
introduced online data augmentation methods,
that help when training with limited data, where the discriminator overfits.
Finally, \cite{karras2021alias} improved the signal flow in the generator.
In our work, we build upon these powerful models in order to achieve multi-modal reflectance generation.

\subsection{Facial Reflectance Acquisition}
\label{sec:related_reflectance}
The first device to accurately acquire 3D facial scans was the LightStage \cite{debevec2000acquiring}, a room-sized dome equipped with programmable illumination and high-end cameras. The diffuse and specular components of the reflectance can be separated 
by exploiting polarization in multiple captures of gradient illumination \cite{ghosh2011multiview}.
Simplified methods have also been proposed, using unpolarized binary patterns \cite{kampouris2018diffuse} or passive illumination \cite{riviere2020single}. Although highly accurate, the above methods require large and expensive equipment.
A recent practical system made of commodity devices \cite{lattas2022practical} enables significantly faster and cheaper facial reflectance acquisition.

Inverse Rendering approaches have also been successful in acquiring facial reflectance. Recent works employ differentiable ray tracing algorithms \cite{dib2021practical, dib2021towards},
to solve an inverse rendering optimization that yields facial diffuse and specular components, as well as specular roughness. However, such methods are computationally expensive and the optimization is susceptible to ambiguity in the subject images.

To overcome such ambiguities, multiple approaches use linear or deep models as priors,
or directly regress the facial reflectance. Early approaches proposed deep image-translation models \cite{chen2019photo,shu2017neural,sengupta2018sfsnet}, while later methods \cite{huynh2018mesoscopic,saito2017photorealistic,yamaguchi2018high} first acquired realistic facial albedos and displacement textures, using one or more deep neural networks.
GANFIT \cite{gecer2019ganfit} was the first method to use a linear shape model in combination with a GAN-based texture model. Nevertheless, the inferred texture contained baked environment illumination, making photorealistic rendering impossible. AvatarMe \cite{lattas2020avatarme} introduced a super-resolution and an image-translation network that transformed textures generated from GANFIT into high-resolution facial reflectance. Its extension AvatarMe++ \cite{lattas2021avatarme++} introduced a generalized facial model that works with arbitrary 3DMM fitting algorithms or scanning methods, by using a differentiable shader. However, both methods are separated from the initial fitting method and cannot pick-up all the facial and illumination cues of the input image.
An alternative approach \cite{bao2021high}
produces highly realistic facial shape and albedo,
but requires a video stream of RGB and depth data for optimization.
Moreover, TRUST \cite{feng2022towards} propose a method to overcome the ambiguity between
skin-tone and illumination optimization,
by using cues from the background illumination or multiple faces in the same image.
Closest to our work,
Luo et al.~\cite{luo2021normalized} acquire normalized avatars using a style-based generator together with an iterative optimization.
Although similar, our approach has the following advantages:
a) our generative model is extended to complete facial reflectance, and our renderer is extended with a more photorealistic shading model,
b) our branched generator and discriminator approach generates additional modalities,
c) our fitting pipeline introduces a latent space and shading regularization, as well as identity loss, and
d) our fine-tuning results in an editable latent space,
which can be further optimized through rendering.

\section{Method}
\label{sec:method}

In this work, we present \ModelName{}, a deep facial reflectance 3DMM,
which is based on a branched multi-modal style-based generative network (Sec.~\ref{sec:method_3dmm}).
It is trained on a big capture-quality facial reflectance dataset,
which is augmented to balance skin tones (Sec.~\ref{sec:method_dataset}). 
The nature of the data allows accurate diffuse and specular differentiable rendering, using an appropriate shader (Sec.~\ref{sec:method_rendering}).
Finally, we combine a style-based generator latent space projection
with 3DMM fitting methods (Sec.~\ref{sec:method_fitting})
and achieve high fidelity facial reconstruction, from a single or multiple images.
The acquired shape and reflectance avatars achieve high identity similarity (given the model's flexibility, diverse dataset and rendering),
have expression blendshapes and can be directly used by common rendering applications.

\subsection{The \ModelName{} Deep 3D Morphable Model}
\label{sec:method_3dmm}

\begin{figure}[h]
  \centering
  \includegraphics[width=\linewidth]{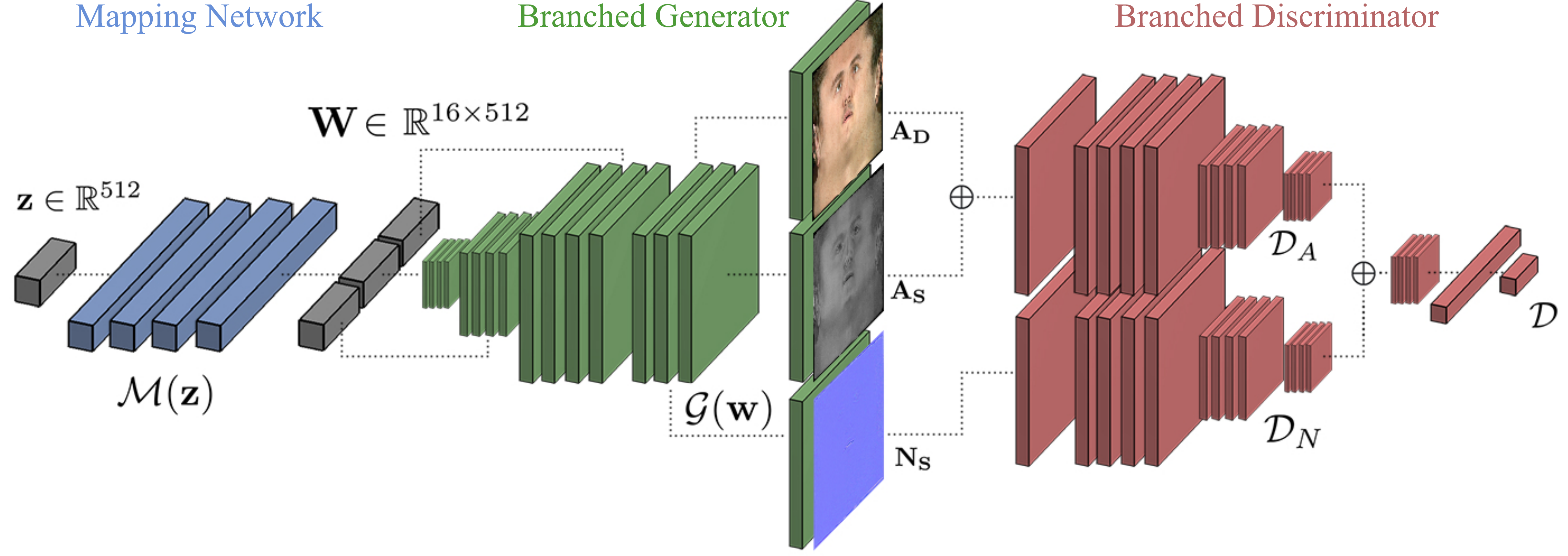}
  \caption{
    Overview of \GANName{}, a style-based generator \cite{karras2020analyzing},
    consisting of a mapping network $\mathcal{M}$ which translates the latent vector $\mathbf{z}$,
    to style $\mathbf{W}$.
    The branched, multi-modal synthesis network $\mathcal{G}$
    generates a diffuse albedo $\mathbf{A}_D$, specular albedo $\mathbf{A}_S$ and normals $\mathbf{N}_S$ at $1024\times1024$ resolution.
    The model is trained in tandem with a a branched multi-modal discriminator $\mathcal{D}$.
    Each branch enables $\mathcal{D}$ to model the distribution of the albedo and normals separately
    The output of each branch is concatenated before entering the last
    convolutional block and the fully connected layers of $\mathcal{D}$,
    in order to maintain feature consistency between modalities.
  }
  \label{fig:method_model}
\end{figure}

We introduce \GANName{},
a multi-modal style-based generative network,
which concurrently generates facial diffuse albedo $\mathbf{A}_D$,
specular albedo $\mathbf{A}_S$ and surface normals $\mathbf{N}_S$
in a UV parameterisation, at $1024\times1024$ resolution.
The model is trained by a novel branched discriminator,
that ensures the consistency between the modalities.
The shape is modeled with a 3DMM and we model both the facial and the head mesh topology similar to \cite{ploumpis2020towards}.

Generating the shape as another branch of \GANName{}, requires many redundant parameters, is prone to quantization,
and its fitting is problematic.
A recent approach generates 3DMM offsets as a UV texture map, but still requires 3DMM projection to generate the final shape \cite{luo2021normalized}.
We opt instead to use a 3DMM with a mesh representation.
We use the Universal Head Model \cite{ploumpis2020towards}, with a facial UV topology, which is completed after fitting.
For a set of identity $\mathbf{p_s}$ and expression $\mathbf{p_e}$ parameters,
with identity and expression bases $\mathbf{U_s}$ and $\mathbf{U_e}$, respectively, and mean $\mathbf{m_s}$,
the facial geometry $\mathbf{S}(\mathbf{p_s, p_e}) \in \mathbb{R}^{106317\times3}$ can be reconstructed as:
\begin{equation}
    \mathbf{S}(\mathbf{p_s, p_e})
    = \mathbf{m_s} 
    + \mathbf{U_s}\mathbf{p_s}
    + \mathbf{U_e}\mathbf{p_e}
    \label{eq:uhm}
\end{equation}

Based on StyleGAN2 \cite{karras2020analyzing},
\GANName{} consists of a mapping network $\mathcal{M}: \mathbf{z} \rightarrow \mathbf{W}$,
which translates a latent vector $\mathbf{z} \in \mathbb{R}^{512}$
into the latent space $\mathbf{W} \in \mathbb{R}^{16\times512}$,
a multi-modal branched synthesis network 
$\mathcal{G}: \mathbf{W} \rightarrow \mathbf{T}$,
where $\mathbf{T}_R=\left\{\mathbf{A}_D, \mathbf{A}_S, \mathbf{N}_S\right\} \in \mathbb{R}^{7\times1024\times1024}$.
Both the diffuse albedo and normals have 3 channels,
however, we use a monochrome specular albedo, which is typical practice in shading \cite{lattas2021avatarme++, kampouris2018diffuse}.
The synthesis network $\mathcal{G}$ follows the skip-connection convolutional blocks architecture
of StyleGAN2ADA\cite{karras2020training},
however, the last block of each up-sampled resolution is branched per reflectance mode,
as shown in Fig.~\ref{fig:method_model}.
For resolutions of $512\times512$ and up, we find this necessary, in order to achieve satisfactory FID scores.
Using a single generator, while branching the final block for each modality,
enables the details to be accurately captured between modalities and ensures the final output corresponds to the same identity.

\GANName{} is also trained using a branched discriminator $\mathcal{D}$.
We observed that the distributions of the albedos $\mathbf{A}_D$, $\mathbf{A}_S$
with mean $0.24$ and std.~$0.12$,
and the normals $\mathbf{N}_S$ with mean $0.61$ and std $0.24$,
were significantly different, something which inhibited the GAN training using a vanilla $\mathcal{D}$ (i.e., a discriminator with no branches).
Therefore, we designed $\mathcal{D}$ as a long-branched network,
based on the residual-based discriminator of StyleGAN2-ADA\cite{karras2020training}.
One branch $\mathcal{D}_A$ receives as input the concatenation 
of the diffuse and specular albedos $\mathbf{A}_D \bigoplus \mathbf{A}_S$,
which have similar statistics,
while the other branch $\mathcal{D}_N$ receives the surface normals $\mathbf{N}_S$.
After running all-but-the-last convolutional blocks,
the output of both branches is concatenated and given to the last convolutional block
and the fully connected layers of $\mathcal{D}$.
This ensures that $\mathcal{D}$ can accurately capture the much different distribution of the albedo and normals,
while ensuring consistent facial features per generated subject.
A detailed visualization is shown in Fig.~\ref{fig:method_overview}.
Luo et al.~\cite{luo2021normalized} recently proposed a similar model, 
where the generator outputs an albedo and a shape,
and is trained by three complete and separate discriminators. 
Our approach, not only requires a smaller network,
but is also able to generate facial reflectance modalities of highly different distributions,
and, as we demonstrate in the experiments, outperforms their generalization capabilities.

\subsection{Dataset Acquisition and Augmentation}
\label{sec:method_dataset}

\begin{figure}[h]
  \centering
  \includegraphics[width=\linewidth]{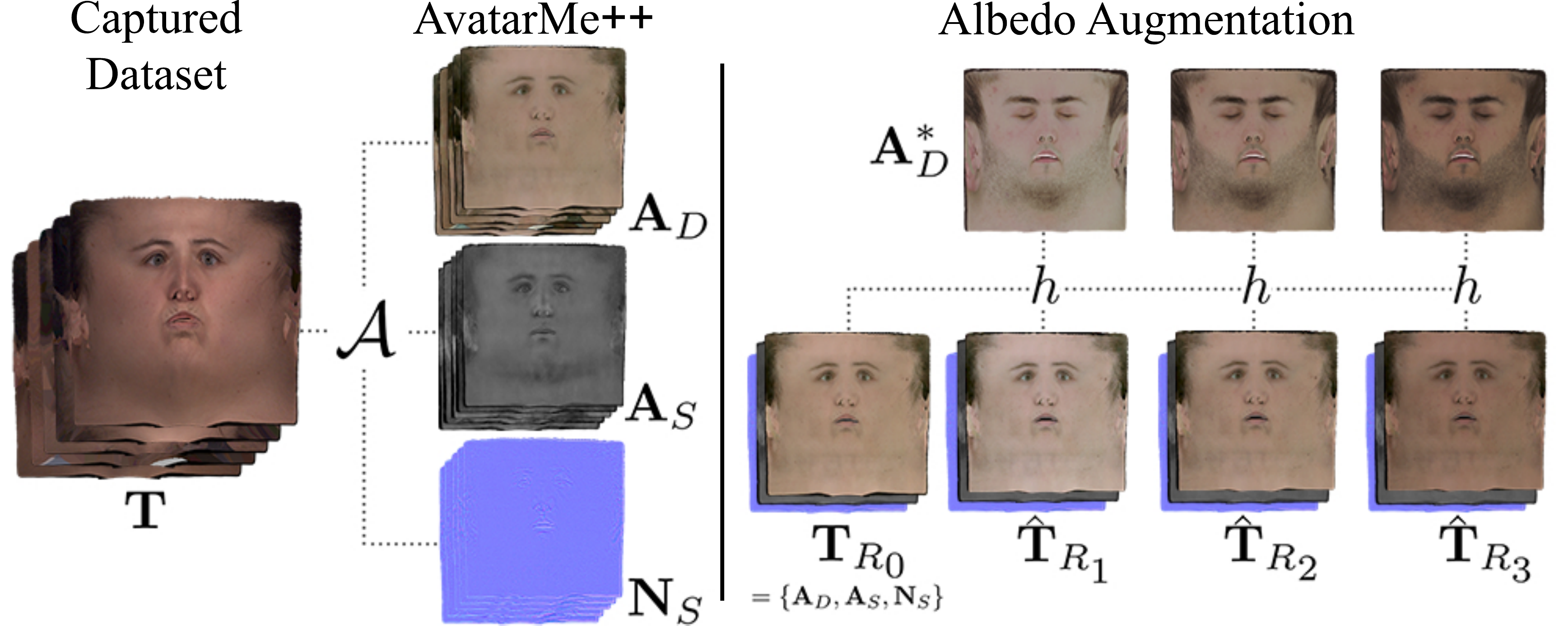}
  \caption{
    Dataset and augmentation.
    Left: creation of diffuse $\mathbf{A}_D$ and specular $\mathbf{A}_S$ albedo and normals $\mathbf{N}_S$,
    using finetuned a AvatarMe++ \cite{lattas2021avatarme++} network $\mathcal{A}$
    on the MimicMe dataset \cite{papaioannou2022mimicme} of textures $\mathbf{T}$.
    Right: albedo skin-tone augmentation using our masked histogram matching $h$ on different target albedos $\mathbf{A}^{*}_D$ (Sec.~\ref{sec:method_dataset}).
  }
  \label{fig:method_augment}
\end{figure}

State-of-the-art 3DMMs and GANs require thousands of data samples 
\cite{booth20163d, karras2020training}, 
however no facial relfectance datasets exist at this scale.
Instead, we utilize a recently published large dataset of facial captures \cite{papaioannou2022mimicme},
and a state-of-the-art facial appearance generation model \cite{lattas2021avatarme++},
to create such a dataset at capture-like quality and high-resolution.
Finally, we alleviate the skin-tone imbalance in the dataset,
by augmenting the resulting data with histogram matching \cite{gonzalez2009digital}.

Firstly, we acquire the MimicMe dataset \cite{papaioannou2022mimicme}, 
consisting of $\mathcal{T} = \{\mathbf{T}_0, \dots, \mathbf{T}_{n_T}\}, n_T = 4,700$ neutral facial textures,
which are projected to a common UV map parameterization.
These textures include diverse subjects and expressions \cite{papaioannou2022mimicme},
however, they have baked illumination from the capturing system and do not constitute an albedo.
We train an image-to-image translation network 
$\mathcal{A}: T \rightarrow \{\mathbf{A}_D, \mathbf{A}_S, \mathbf{N}_S\}$
that transforms textures from $\mathcal{T}$
to reflectance textures $\{\mathbf{A}_D, \mathbf{A}_S, \mathbf{N}_S\}$, following AvatarMe++ \cite{lattas2021avatarme++},
while making the following changes:
(a) we approximate the capturing environment of \cite{papaioannou2022mimicme} using AvatarMe++ \cite{lattas2021avatarme++},
and train the AvatarMe++ network $\mathcal{A}$ for this environment only,
(b) we generate only the surface (i.e., specular) normals $\mathbf{N}_S$,
in tangent space, which aids the \ModelName{} fitting, as discussed in \ref{sec:method_rendering} and
(c) we do not use a super-resolution network, since the textures in $\mathcal{T}$ are high-resolution captured textures.
In this manner, we can use $\mathcal{A}$ to transform $\mathcal{T}$ 
into a dataset of reflectance textures $\mathcal{T}_R$.

Regarding the shape, we choose not to represent it as a UV map,
and instead we use an existing PCA model.
Although we can extract the shape from the above datasets,
our experiments show no improvement over using PCA,
while also making the network training more challenging.
Therefore, with our method being shape-model agnostic,
we use the public UHM head shape model \cite{ploumpis2020towards}.

Biased skin tone prediction is a significant problem \cite{feng2022towards},
and therefore balanced skin tone representation in the dataset is of paramount importance.
Skin tone augmentation can be done by manipulating the melanin and hemoglobin values of the albedo,
which requires expensive data collection \cite{gitlina2020practical}, pre-training \cite{aliaga2022estimation} or potentially noisy inverse rendering \cite{alotaibi2017biophysical}.
Instead, we propose an alternative method based on histogram matching \cite{gonzalez2009digital} (Overview in Fig.~\ref{fig:method_augment}),
which is trivial to use and noise-free.
We acquire 10 scanned albedos \cite{ghosh2011multiview},
with each subject matched to the Monk Skin-Tone Scale (MST) \cite{Monk2022MST}
and transform them to our facial UV topology.
Then, for each albedo $\mathbf{A}_D$ in the reflectance dataset, 
we perform histogram matching $h$ \cite{gonzalez2009digital}
with a number of randomly selected target albedos $\mathbf{A}_D^{*}$ from the MST scale albedos.
To avoid non-skin features that could affect the transformation,
we calculate a mask $\mathbf{M}$ using the distance of each pixel from the average skin tone $\bar{\mathbf{A}_D}$ 
in the forehead region $\mathbf{M} = \left|\left| \mathbf{A}_D - \bar{\mathbf{A}_D}\right|\right|$. Then,
we acquire an augmented albedo $\hat{\mathbf{A}_D}$ as:
\begin{equation}
  \hat{\mathbf{A}_D} = \mathbf{M}\odot
                h(\mathbf{A}_D, \mathbf{A}_D^{*}) 
                + (1-\mathbf{M})\odot\mathbf{A}_D
  \label{eq:method_histogram}
\end{equation}

\subsection{Fast Differentiable Photorealistic Rendering}
\label{sec:method_rendering}

Typically, 3DMM models rely on a single RGB per-vertex texture color,
a pinhole camera projection model and a diffuse-only lambertian shading model,
based on spherical harmonics illumination \cite{egger20203d}.
Although such modeling provides enough cues to the fitting function
for the shape and texture parameters to be optimized,
the texture model cannot capture the facial reflectance parameters required for photorealistic rendering and the lambertian shading model cannot handle both the diffuse and specular components.

On the contrary, differentiable local-reflection models such as Blinn-Phong \cite{blinn1977models}
can achieve more accurate facial shading \cite{lattas2021avatarme++}.
We define a viewing direction $\mathbf{v}$, an ambient intensity $\mathbf{c_a}$, and a set of $n_l$ light sources with direction $\mathbf{l_j}$ and intensity $\mathbf{c_j}$.
For a rasterized pixel $i$ 
with diffuse albedo $\mathbf{A}_D$, diffuse normals $\mathbf{N}_D$,
specular albedo $\mathbf{S}_A$ and surface normals $\mathbf{N}_S$,
the diffuse $\mathbf{U_{i_D}}$ and specular $\mathbf{U_{i_S}}$ shading components are defined as:
\begin{equation}
    \mathbf{U_{D_i}} = 
        \mathbf{c_a}
        \mathbf{A_{D_i}}
        \sum_{j = 1}^{n_l}
        (
            \mathbf{N_{D_i}} \cdot \mathbf{l_j}
        ) \mathbf{c_j}
\label{eq:background_shading_diffuse}
\end{equation}\vspace{-0.5cm}
\begin{equation}
    \mathbf{U_{S_i}} = 
        \mathbf{A_{S_i}}
        \sum_{j = 1}^{n_l}
        (
            \chi^{+}
            (
                \mathbf{N_{S_i}} \cdot \mathbf{h_j}
            )
        )^{s} \mathbf{c_j}
        , \quad
        \mathbf{h_j}
        =
        \frac{
            \mathbf{l_j+v_j}
        }{
            \left|\left|\mathbf{l_j+v_j}\right|\right|
        }
\label{eq:background_shading_specular}
\end{equation}
where $\chi^{+}(x)$ is a piece-wise function that returns $\max{\{0, x\}}$, and $s$ is the shininess coefficient.
The final shading is acquired by addition 
$\mathbf{U_{i}} = \mathbf{U_{i_D}} + \mathbf{U_{i_S}}$.
We also perform subsurface scattering using \cite{lattas2021avatarme++}.

For photorealistic shading, separate diffuse and surface normals are required \cite{lattas2021avatarme++,ghosh2011multiview,ma2007rapid},
so that the diffuse component approximates the smoother subsurface scattering.
In order to correctly optimize the shape mesh during fitting,
we use the smoother geometric normals as diffuse normals $\mathbf{N}_D$.
The detailed specular normals
are generated by \GANName{} in tangent space $\mathbf{N}_S$,
and during rendering are added to the shape normals $\mathbf{N}$,
to acquire the specular normals in object space.

Our differentiable rendering implementation,
not only renders skin photorealistically compared to recent fitting methods \cite{gecer2019ganfit, luo2021normalized, smith2020morphable},
but it is also implemented directly in image-space, compared to \cite{lattas2021avatarme++},
so that direct supervision can be achieved.
Moreover, the 3DMM shape $\mathbf{S}$
and GAN-generated reflectance texture maps $\mathbf{T}_R = \{ \mathbf{A}_D,\mathbf{A}_S,\mathbf{N}_S\}$ can be efficiently optimized by photometric, identity and perceptual losses.
The rendering function $\mathcal{R}$ with optimizable camera $\mathbf{p}_c$ and illumination $\mathbf{p}_l$
can be formulated as:
\begin{equation}
\mathbf{I}_\mathcal{R} = \mathcal{R}\left(\mathbf{S}, \mathbf{T}_R, \mathbf{p_c}, \mathbf{p_l}\right)
\end{equation}

\subsection{Fitting by Inversion through Rendering}
\label{sec:method_fitting}

Following conventional 3DMM fitting approaches \cite{blanz1999morphable,romdhani2005estimating,egger20203d,booth20163d,gecer2019ganfit}, we build an optimization pipeline with two major improvements: 
(a) we replace the statistical texture model and its PCA-based optimization,
with \GANName{} textures and GAN inversion optimization \cite{karras2020analyzing, roich2022pivotal},
b) we implement an accurate differentiable renderer in the image-space,
to enable \GANName{} inversion, by rendering its results in high-fidelity.
As described below, our optimization pipeline combines the merits of both
3DMM fitting and GAN inversion and achieves photorealistic reflectance reconstruction.

\subsubsection{\GANName{} Inversion}
Generator inversion refers to the acquisition of latent code 
that recreates the target image
and its application on style-based generators is a widely studied problem \cite{karras2020analyzing, roich2022pivotal}.
However, all the above assume a 2D target image and can directly optimize the generator's output on it.
Moreover, \ModelName{} requires to pass the generated reflectance through rasterization and rendering, 
while preserving the meaningful properties of the texture map,
for which a standard inversion does not suffice.
Inspired by both previous 3DMM fitting methods \cite{gecer2019ganfit, lattas2021avatarme++, booth20163d, dib2021practical},
and inversion \cite{karras2020training, roich2022pivotal},
we design a rendering-based inversion method for our multi-modal generator.

Style-based generators \cite{karras2020analyzing,karras2020training} feed a latent code $\mathbf{z}$ into a mapping network, to acquire the native latent space $\mathbf{W}$, which is fed to the synthesis network.
Inversion methods optimize $\mathbf{W}$ directly,
whilst also optimizing the noise vector $\mathbf{n}$ \cite{karras2020training, roich2022pivotal},
by using a perceptual LPIPS loss \cite{zhang2018unreasonable}.
However, the multi-modal generator faces three challenges:
a) being multi-modal makes the concurrent optimization of the different modes more difficult,
b) requires rasterization and as such not all pixels are visible to the loss function and 
c) requires shading, shape, camera and light source optimization,
which all remain sensitive to the optimization with LPIPS.
We find that an additional set of losses make the inversion more accurate and stable:

The terms of our objective function includes both primitive supervision such as landmark, photometric, and regularization loss functions, as well as identity and perceptual losses.
For a target image $\mathbf{I_0}$ and a rendered reconstruction $\mathbf{I}_{\mathcal{R}}$,
we used the following losses to optimize the shape 3DMM, the camera and illumination parameters,
and also to guide the rasterized \GANName{} results:

\noindent{\textbf{Landmark Loss:}} We estimate 3D landmarks using a deep alignment network~\cite{deng2020retinaface} $\mathcal{M}(\mathbf{I}): \mathbb{R}^{H\times W \times C} \rightarrow \mathbb{R}^{68 \times 2}$ 
and penalize the distance by $ \mathcal{L}_{lan} = ||\mathcal{M}(\mathbf{I_0}) -  \mathcal{M}(\mathbf{I}_{\mathcal{R}})||_2$.

\noindent{\textbf{Photometric Loss:}} It is defined as the distance between per-pixel color intensity values,
to capture the skin color and illumination from the target image as $\mathcal{L}_{ph} = ||\mathbf{I_0} - \mathbf{I}_{\mathcal{R}}||_1 $.

\begin{figure*}[ht]
  \centering
  \includegraphics[width=\linewidth]{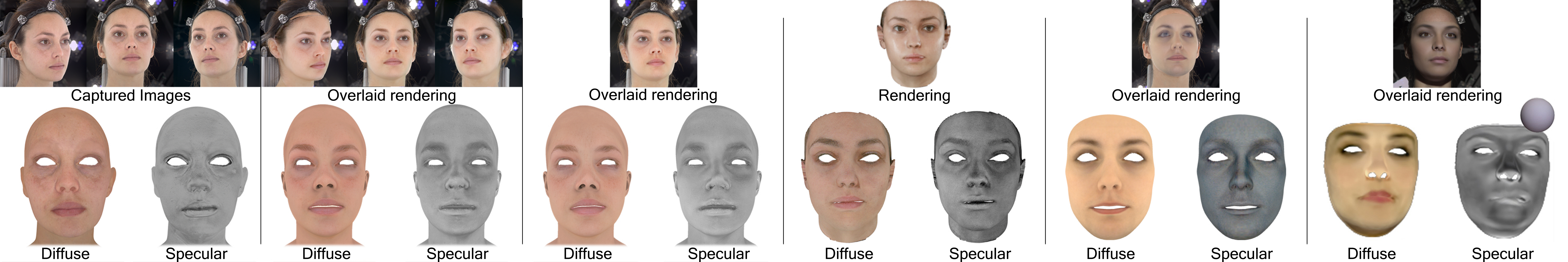}
    \makebox[0.16\linewidth][c]{\footnotesize{Ground Truth \cite{alexander2010digital}}}\hfill
    \makebox[0.16\linewidth][c]{\footnotesize{Ours, 3 input images}}\hfill
    \makebox[0.16\linewidth][c]{\footnotesize{Ours, 1 input image}}\hfill
    \makebox[0.16\linewidth][c]{\footnotesize{AvatarMe++\cite{lattas2021avatarme++}}}\hfill
    \makebox[0.16\linewidth][c]{\footnotesize{AlbedoMM \cite{smith2020morphable}}}\hfill
    \makebox[0.16\linewidth][c]{\footnotesize{Dib et al. 2021~\cite{dib2021towards}}}\hfill
    \vspace{-0.2cm}
  \caption{
    Comparison of diffuse and specular albedo reconstruction and rendering,
    of Digital Emily \cite{alexander2010digital} with prior works.
    Both our single and three images reconstruction,
    achieve similar results to the captured data,
    which need specialized hardware and hundreds of images.
  }
  \label{fig:exp_emily}
  \vspace{-0.4cm}
\end{figure*}

\noindent{\textbf{Identity Loss:}} Following~\cite{gecer2021fast,genova2018unsupervised}, we supervise fitting by identity features extracted by a face recognition network~\cite{deng2019arcface} with $n$ layers, $\mathcal{F}^n(\mathbf{I}): \mathbb{R}^{H\times W \times C} \rightarrow \mathbb{R}^{512}$ to provide strong identity similarity to the target image:
\begin{equation}
\mathcal{L}_{ID} = 1 - \frac{\mathcal{F}^n( \mathbf{I_0}) \cdot\mathcal{F}^n( \mathbf{I}_\mathcal{R})}{||\mathcal{F}^n( \mathbf{I_0})||_2 \cdot||\mathcal{F}^n(\mathbf{I}_\mathcal{R})||_2}
\label{eq:id_loss}
\end{equation}
Along with the abstract level supervision provided by identity features, we also optimize for intermediate activations of the face recognition network to ensure mid-level perceptual information can be reconstructed as well:
\begin{equation}
    \mathcal{L}_{per} =  \sum_j^n{\dfrac{||\mathcal{F}^j(\mathbf{I_0}) - \mathcal{F}^j(\mathbf{I}_{\mathcal{R}})||_2}{H_{\mathcal{F}^j} \cdot W_{\mathcal{F}^j} \cdot C_{\mathcal{F}^j}}}
\end{equation}

\noindent{\textbf{{\GANName{} Regularization:}}}
Rasterization results in various pixels and facial areas not being used during optimization, while shading may be imperfect in complex illumination environments,
both of which introduce noise and environmental features to our albedo and normals during $\mathbf{W}$'s optimization.
To avoid such artifacts, we follow the $\mathbf{W}$ initialization protocol introduced in \cite{karras2019style}, and on top we apply an $L_2$ constraint to make sure $\mathbf{W}$'s values do not greatly deviate from their feasible space. 

\noindent{\textbf{Shape Regularization:}} We follow the literature~\cite{booth20173d} in
constraining the shape parameters weighted by their inverse eigenvalues:
$\mathcal{L}_{s} = \left|\left| \mathbf{p}_s\right|\right|_{\mathbf{\Sigma}_s^{-1}}^2$
and 
$\mathcal{L}_{e} = \left|\left| \mathbf{p}_e\right|\right|_{\mathbf{\Sigma}_e^{-1}}^2$,
where $\mathbf{\Sigma}$ denotes a diagonal matrix with the eigenvalues.

\noindent{\textbf{Overall loss:}}
It is defined as follows, where all $\lambda$'s are hyper-parameters suitably chosen to balance the different loss terms during optimization:
\begin{align}
    \mathcal{L} = &\lambda_{1}\mathcal{L}_{lan} + \lambda_{2}\mathcal{L}_{ph} + \lambda_{3}\mathcal{L}_{ID} + \lambda_{4}\mathcal{L}_{per}\nonumber\\
        &+ \lambda_{5}\mathcal{L}_\mathbf{W} + \lambda_{6}\mathcal{L}_{s} + \lambda_{7}\mathcal{L}_{e},
    \label{eq:inversion_loss}
\end{align}
\subsubsection{\GANName{} Tuning}
Recent works show that $\mathbf{W}$-based inversion cannot fully recover the target image
\cite{roich2022pivotal}.
One approach is to optimize the extended latent space $\mathbf{W}+$ \cite{abdal2021styleflow, luo2021normalized},
however this inhibits latent code manipulations \cite{roich2022pivotal},
while, in our experiments, ends up in an outright copy of the target image into the albedo,
inhibiting also the re-rendering of the avatar.
On the other hand, Pivotal Tuning Inversion (PTI) \cite{roich2022pivotal},
fine-tunes the generator on the target image, after finding and freezing $\mathbf{W}$.
We find this approach much more accurate in our case,
as only the visible texture parts produce gradients and are optimized.
Again, given imperfect illumination optimization and shape fitting,
PTI transfers high-frequency noise, illumination and non-skin features on the albedo,
when tuning the whole generator.

Our experiments show that during fine-tuning, 
the first layers of \GANName{} change the color of the albedo but absorb illumination,
the middle layers change the mesostructure of the skin,
while the last layers change fine details, but absorb noise from the target image.
Therefore, we found that it is optimal to only tune the 4 middle layers,
while keeping the rest frozen.
We use the proposed LPIPS $\mathcal{L}_{LP}$ \cite{zhang2018unreasonable},
using a pre-trained VGG network \cite{simonyan2014very},
and photometric $\mathcal{L}_{ph}$ losses. Moreover, we add two regularization losses often used in inverse rendering \cite{dib2021towards}: i) a horizontal flip loss $\mathcal{L}_{flip}$,
and ii) a chromaticity loss $\mathcal{L}_{\kappa}$,
in order restrict the fine-tuning to add shadows and highlights,
that are not reproduced by the rendering,
to the albedo:
\begin{equation}
    \mathcal{L}_{PTI} = \lambda_{8}\mathcal{L}_{LP} + \lambda_{9}\mathcal{L}_{ph} + 
    \lambda_{10}\mathcal{L}_{flip} + \lambda_{11}\mathcal{L}_{\kappa}
    \label{eq:pti_loss}
\end{equation}
where $\lambda$'s are hyper-params chosen before the fine-tuning.

\section{Experiments}
\label{sec:exp}

\subsection{Implementation Details}
\label{sec:exp_implementation}

Our generator code builds on the public repository of StyleGAN2-ADA \cite{karras2020training}.
The branches follow the same architecture, and are connected through concatenation.
For the differentiable rendering, we use a blinn-phong shader on Pytorch3D \cite{ravi2020pytorch3d} following \cite{lattas2021avatarme++}.
Our fitting code requires on average 50 seconds on a machine with one NVIDIA 2080 GPU.
We run the fitting for 200 iterations for inversion (Sec.~\ref{sec:method_fitting})
and 20 iterations for tuning (Sec.~\ref{sec:method_fitting}).
Detailed loss parameters are included in the supplemental.

\subsection{Single-Image Reconstruction}
\label{sec:exp_single}

\begin{figure}[ht]
  \centering
  \includegraphics[width=0.49\linewidth]{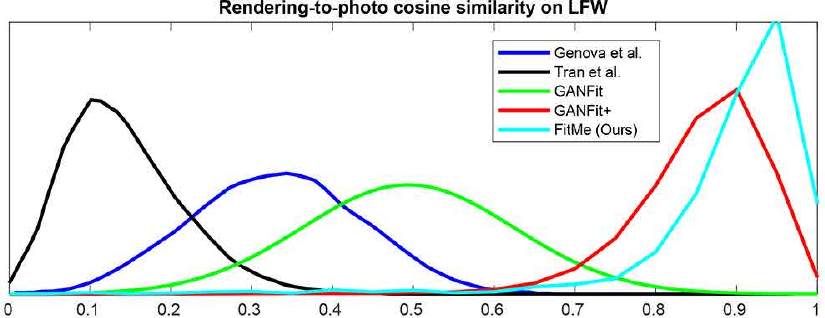}
  \includegraphics[width=0.49\linewidth]{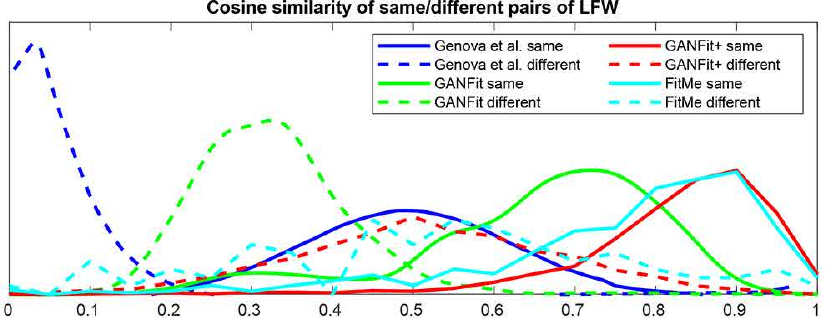}
  \caption{
    Left: Cosine similarity distributions between real image and rendered reconstruction of LFW \cite{LFWTech}, using VGG-Face \cite{parkhi2015a}.
    Right: Cosine similarity distribution between same and different
    pairs of reconstructions.
    Compared with \cite{Genova_2018_CVPR, tuan2017regressing, gecer2019ganfit, gecer2021fast}.
  }
  \label{fig:exp_lfw_similarity}
\end{figure}

\begin{figure}[]
  \centering
  \includegraphics[width=\linewidth]{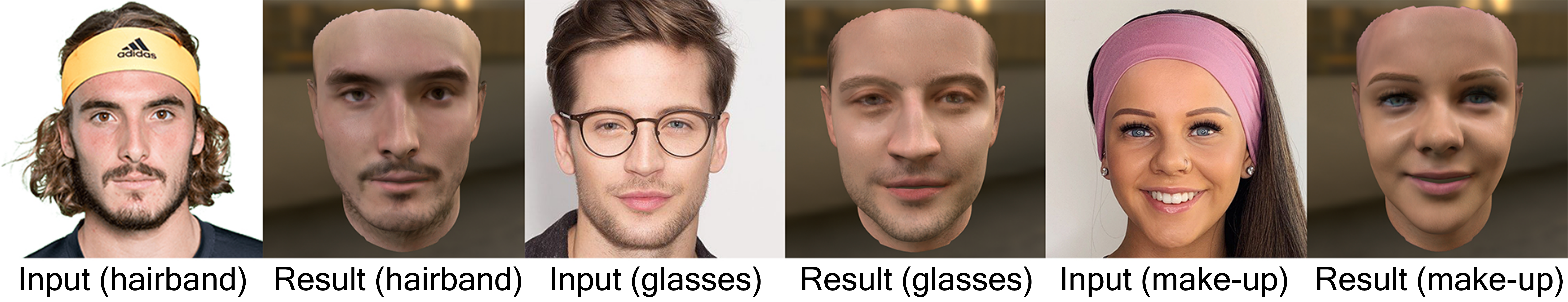}
  \vspace{-0.5cm}
  \caption{
    Results of our method on challenging examples.
  }
  \vspace{-0.3cm}
  \label{fig:challenging}
\end{figure}

\begin{figure}[h]
  \centering
  \includegraphics[width=\linewidth]{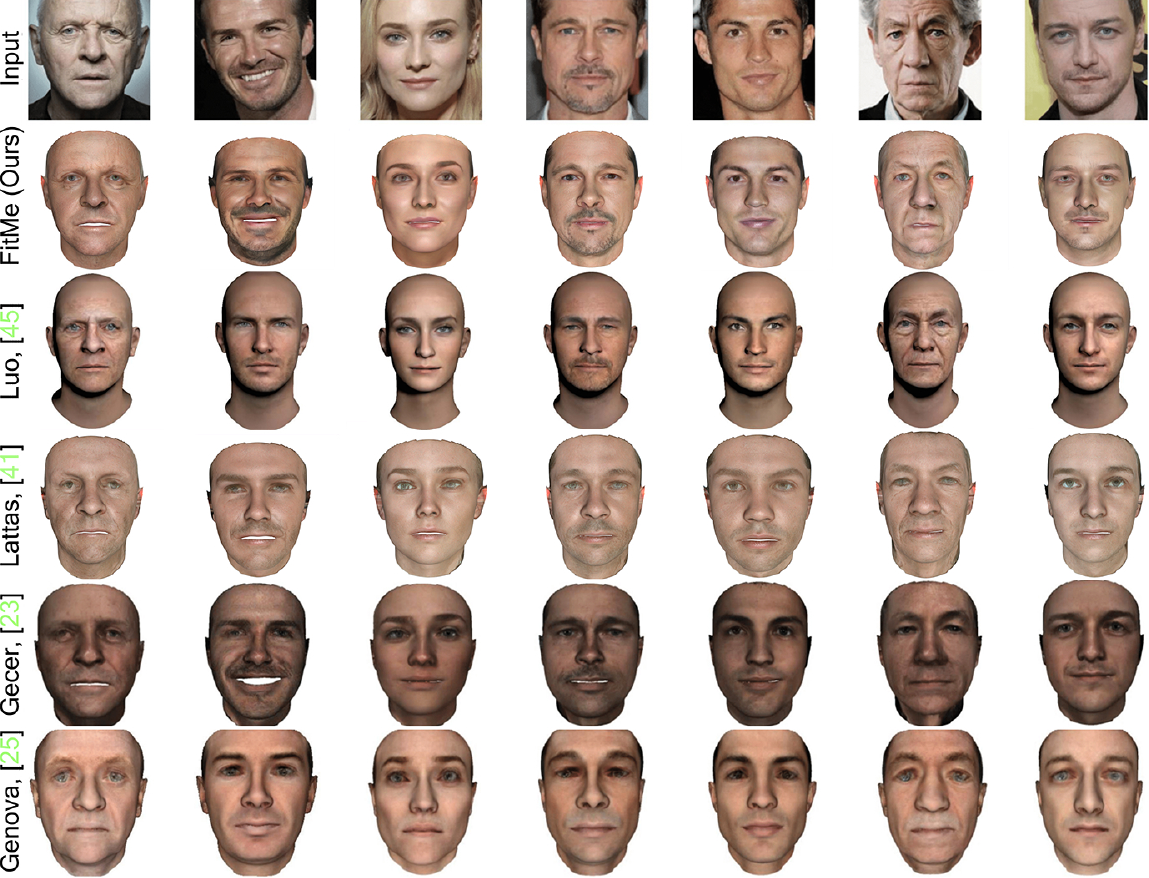}
  \caption{
    Comparison with single-image reconstruction methods \cite{luo2021normalized,lattas2021avatarme++,gecer2019ganfit,genova2018unsupervised}
    Rows 3,5,6 acquired from \cite{luo2021normalized} benchmark.
    \ModelName{} achieves higher likeness and acquires relightable reconstructions.
  }
  \label{fig:comp_luo_gecer_genova}
\end{figure}

\begin{figure}[h]
  \centering
  \includegraphics[width=\linewidth]{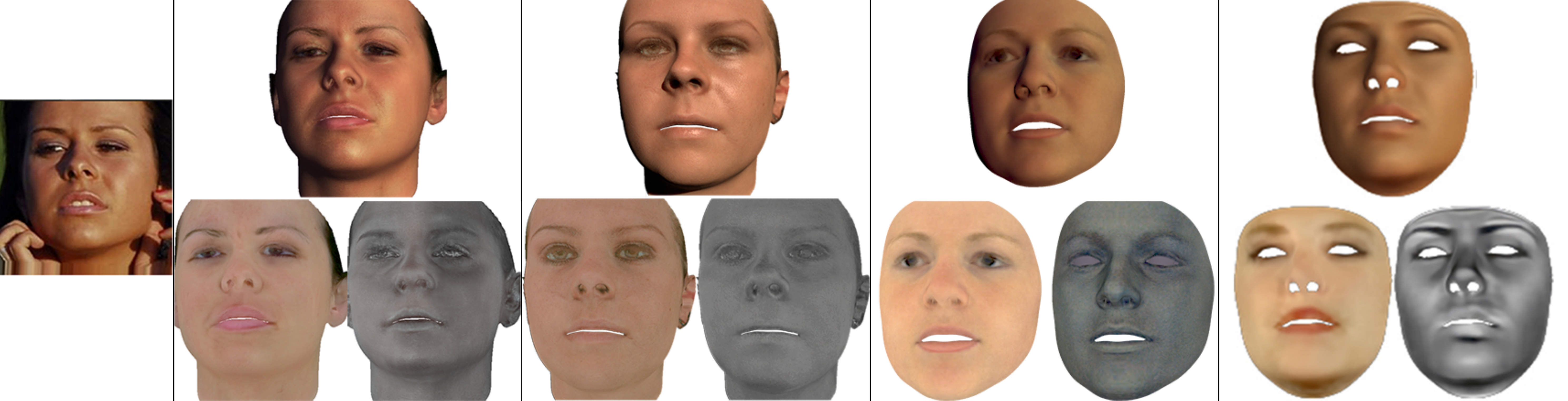}
\makebox[0.11\linewidth][c]{\footnotesize{Input}}\hfill
\makebox[0.22\linewidth][c]{\footnotesize{Ours}}\hfill
\makebox[0.22\linewidth][c]{\footnotesize{AvatarMe\cite{lattas2021avatarme++}}}\hfill
\makebox[0.22\linewidth][c]{\footnotesize{AlbedoMM\cite{smith2020morphable}}}\hfill
\makebox[0.22\linewidth][c]{\footnotesize{Dib et al.~\cite{dib2021towards}}}\hfill
  \caption{
    Comparison with single-image reflectance acquisition \cite{lattas2021avatarme++, smith2020morphable, dib2021towards}.
    Top: rendering, bottom: diffuse and specular albedo.
  }
  \label{fig:comp_dib_avatarme_smtih}
\end{figure}

We evaluate the identity preservation of our single-image reconstruction,
by comparing our results in the LFW dataset \cite{LFWTech} with previous literature
\cite{gecer2019ganfit, gecer2021fast, Genova_2018_CVPR, tuan2017regressing}.
Using a pre-trained face recognition network \cite{parkhi2015a},
we measure the cosine similarity between real image and rendering,
and plot the results in Fig.~\ref{fig:exp_lfw_similarity},
which clearly illustrates the effectiveness of our method.
We additionally plot the cosine similarity of between 
reconstruction pairs of the same identity versus
different identity.
Moreover, in Fig.~\ref{fig:comp_luo_gecer_genova},
we show a qualitative comparison with prior work on single-image reconstruction and in Fig.~\ref{fig:comp_dib_avatarme_smtih},
with single-image reflectance acquisition methods \cite{dib2021practical,lattas2020avatarme,smith2020morphable,lattas2021avatarme++}.
Finally, in Fig.~\ref{fig:challenging} we show our results on challenging examples.
All the above show our method's ability to preserve the identity of the subject
and produce high-quality shape and reflectance.

\begin{figure}[h]
    \centering
    \includegraphics[width=\linewidth]{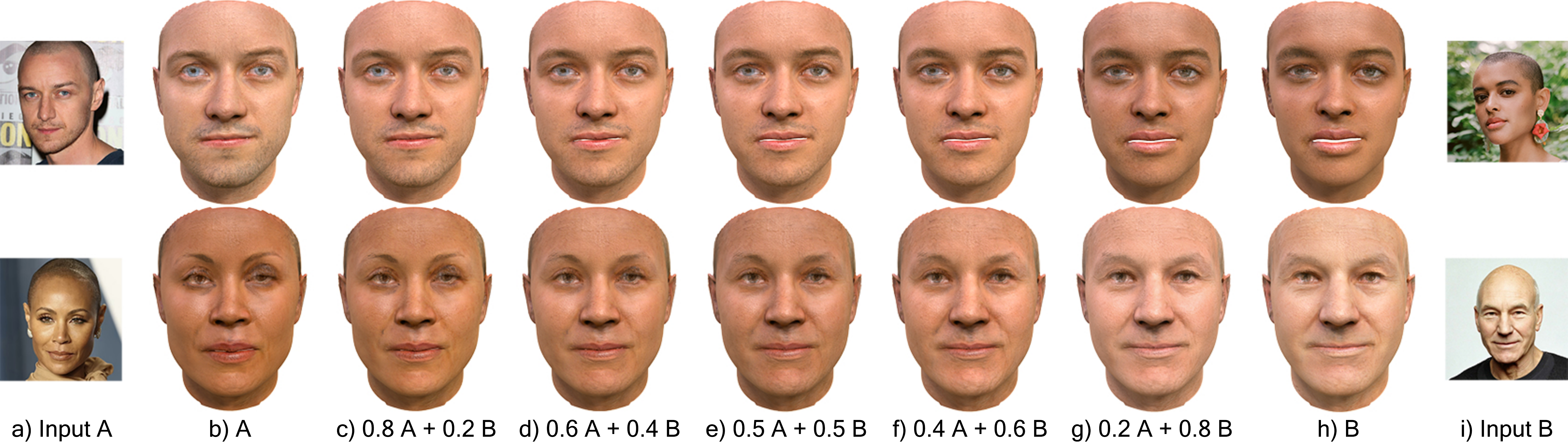}
    \caption{
        Interpolation of the latent parameters between fittings.
        For each row, 
        we interpolate the latent $\mathbf{W}$, 
        and shape parameters $\mathbf{p}_s$, $\mathbf{p}_e$,
        between the left-most (A) and the right-most fitting (B).
    }
    \label{fig:interp}
\end{figure}

Additionally, to show the capabilities of the learned latent space,
we perform interpolations between different fittings, which we visualize in Fig.~\ref{fig:interp}, by linearly blending their parameters.
Our model is able to transition smoothly between fittings, even after generator tuning.
In supplemental, we also show PCA-based editing of the latent space.

\subsection{Multi-Image Reconstruction}
\label{sec:exp_multi}
So far, we have been assuming a single target image,
however, our method can be used as is for multi-view reconstruction.
To do so, we optimize along a batch of $N$ images $\mathbf{I}_{0_i}, i = 0,1,\dots,N$.
For each batch, we separately optimize the camera $\mathbf{p}_{e_i}$,
and illumination $\mathbf{p}_{l_i}$ parameters,
while optimizing a single latent vector $\mathbf{W}$,
and shape vector $\mathbf{p}_s$.
The expression vector $\mathbf{p}_e$ can be varied if needed.
Moreover, we only use one instance of the generator $\mathcal{G}$.
During optimization, we average the loss across the batch.

We find that a frontal and two side images produce high quality
reconstructions that resemble facial capture.
Fig.~\ref{fig:exp_emily} shows a comparison of our 3-image reconstruction and our 1-image reconstruction, with a Light Stage captured Digital Emily \cite{alexander2010digital},
and prior work \cite{lattas2021avatarme++, smith2020morphable, dib2021towards}.
As can be seen, our method can successfully be used for fast shape and reflectance acquisition from multi-view sets.
Finally, Fig.~\ref{fig:teaser} and multiple supplemental figures,
show examples of our reconstruction,
using unconstrained mobile phone images, enabling quick avatar generation.

\subsection{Dataset Augmentation}
\label{sec:exp_augmentation}

Our dataset augmentation achieves comparable performance to physics-based skin models
\cite{gitlina2020practical, aliaga2022estimation}, without requiring the calculation of such complex interactions. 
Moreover, compared to the more practical method \cite{alotaibi2017biophysical},
we avoid inverse-rendering based noise and achieve similar results in $2.5$ seconds per $1024\times1024$ albedo map, compared to the $46.1$ seconds of \cite{alotaibi2017biophysical}.
We provide a comparison in the supplemental materials.

\section{Limitations}
Our method is limited in certain aspects 
pertaining the data and the fitting process.
The data limitations include a) the imbalance of features in the training data,
which we attempt to overcome by the proposed augmentation and generator tuning.
Moreover, the reflectance data are captured with the assumption of skin-like materials
and therefore the eyes exhibit noisy reflections.
Finally, there exists ambiguity between skin-tone and illumination intensity during fitting,
which could be alleviated by combining our method 
with the findings of TRUST \cite{feng2022towards} in future work.

\section{Conclusion}
In this paper we introduced FitMe, a method that is able to produce highly accurate, renderable human avatars based on a single or multiple ``in-the-wild'' images. To achieve this, we introduced a deep facial reflectance model that consists of a multi-branched style-based GAN and a PCA-based shape model, as well as an easy skin-tone augmentation method. Moreover, we presented a novel iterative optimization procedure that is based on a differential rendering process. As we showed in a series of experiments, our method helps to bridge the uncanny valley and creates pleasing results, directly renderable in common renderers.

\section*{Acknowledgments}
AL and SZ were supported by EPSRC Project DEFORM
(EP/S010203/1) and SM by an Imperial College DTA.
\newpage

{\small
\bibliographystyle{ieee_fullname}
\bibliography{egbib}
}


\clearpage
\appendix

\twocolumn[{%
\renewcommand\twocolumn[1][]{#1}%
\begin{center}

\vspace{3pt}
\textbf{{\Large \ModelName{}: Deep Photorealistic 3D Morphable Model Avatars}}\\
\textbf{\Large (Supplementary Material)}
\vspace{30pt}
\end{center}%
}]

\section{Model Manipulation}
\label{sec:model}

\begin{figure}[h]
    \centering
    \includegraphics[width=\linewidth]{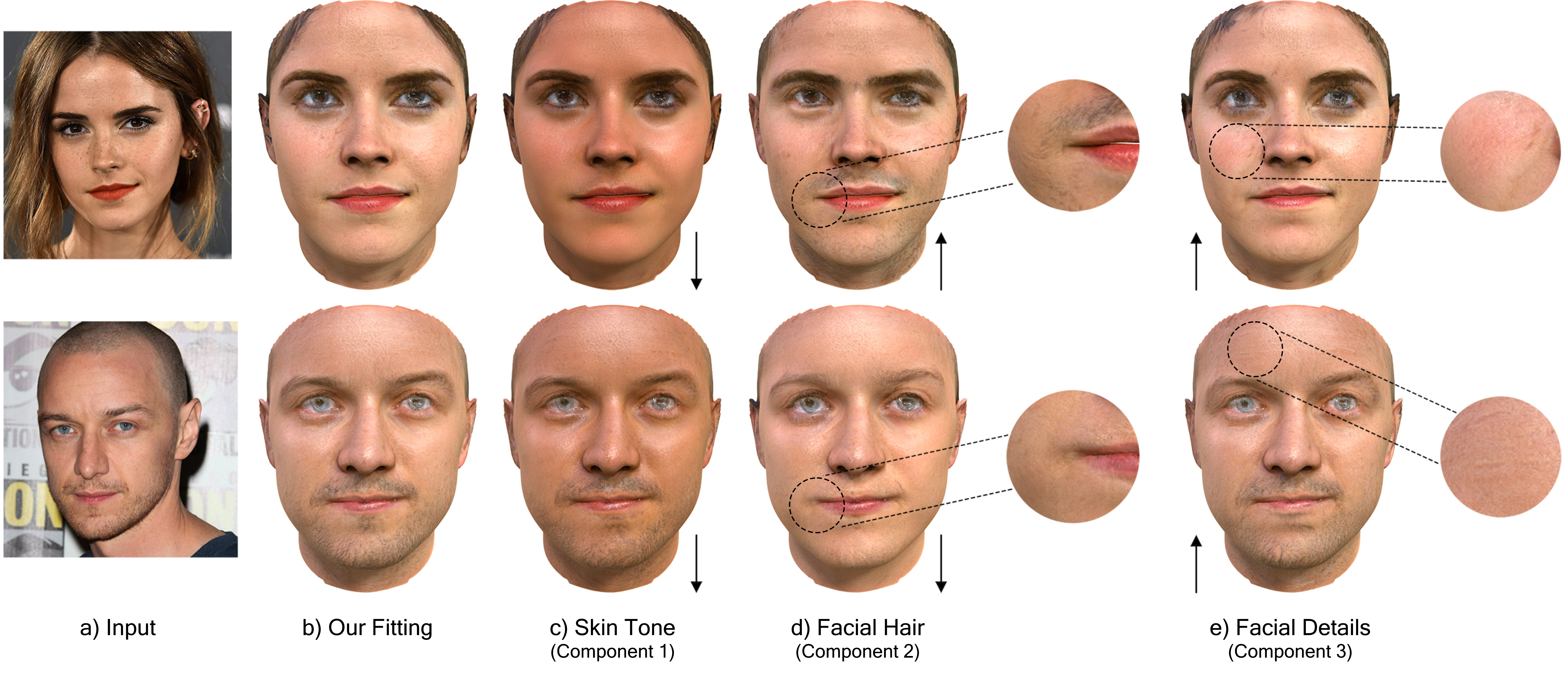}
    \caption{
        Manipulation of a \GANName{} projection.
        By performing PCA on the latent space of \GANName{},
        we can manipulate the most important components,
        which roughly correspond to c) skin tone, d) gender and e) details (wrinkles, freckles).
        The arrows indicate the manipulation direction.
    }
    \label{fig:pca_manip}
\end{figure}

Given enough samples from the trained \GANName{} generator (10000 in our case), the latent space $\mathbf{W}$ of a style-based generator,
can be analyzed with PCA \cite{harkonen2020ganspace}.
We expect the first principal components to expose interpretable controls
over features of the reconstruction.
In Fig.~\ref{fig:pca_manip} we show how the first three components
correspond roughly to the skin tone (given our augmentation),
gender and age variations.
These show that our model learns a meaningful latent space
and enable us to perform semantic manipulations directly on the reflectance UV maps.

\section{Dataset Augmentation}

\begin{figure}[h!]
  \centering
  \includegraphics[width=\linewidth]{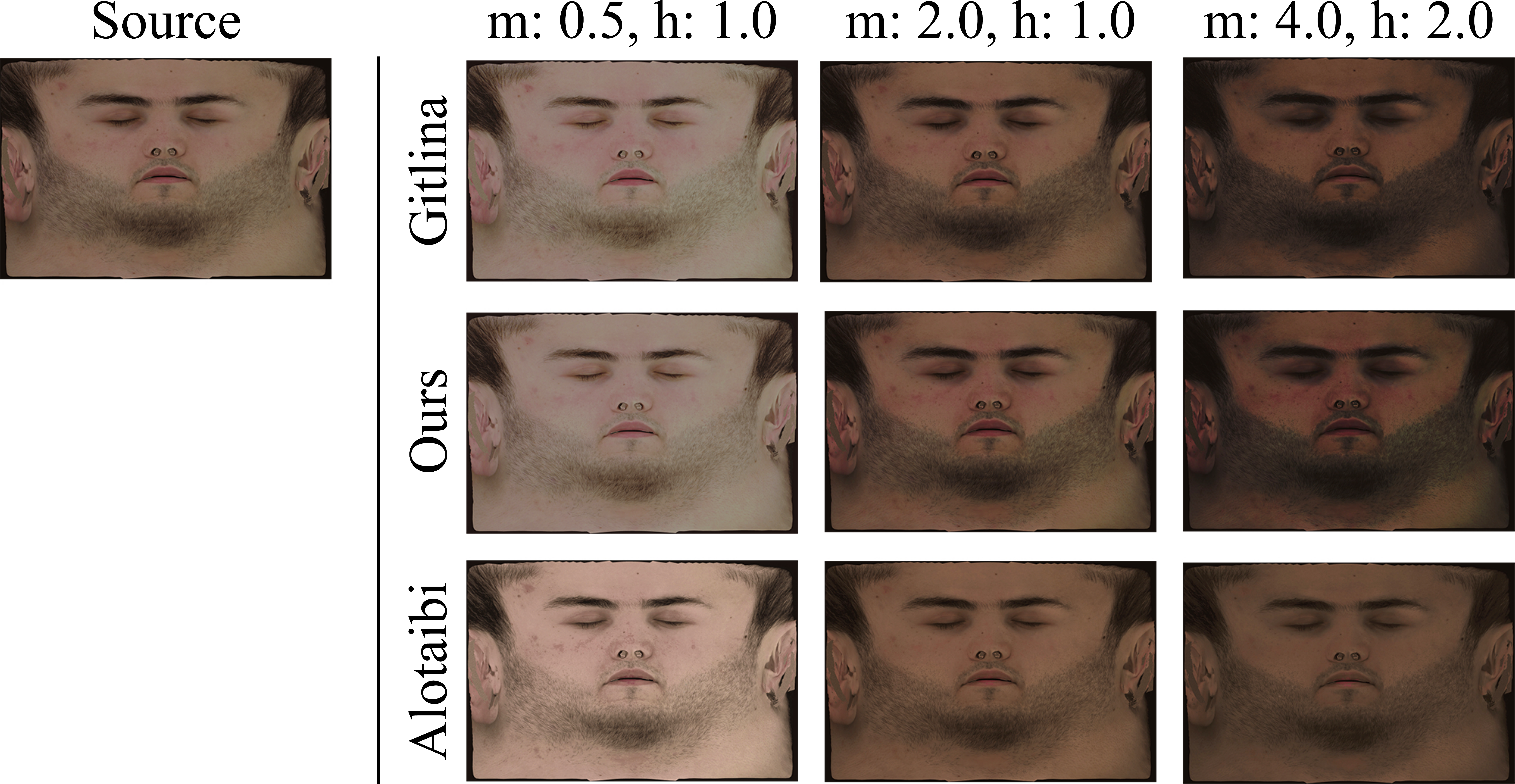}
  \caption{
    Skin tone augmentation comparisons, with adjusted melanin $m$ and hemoglobin $h$, between
    (top) LightStage-measured look-up table melanin-hemoglobin of Gitlina et al.~\cite{gitlina2020practical},
    (middle) our proposed masked histogram matching and
    (bottom) inverse-rendering based manipulation of Alotaibi and Smith~\cite{smith2020morphable}.
  }
  \label{fig:exp_comparison_augment}
\end{figure}

We perform a comparison of our proposed histogram matching albedo augmentation,
against a LightStage captured melanin-hemoglobin manipulation method \cite{gitlina2020practical},
and a melanin-hemoglobin manipulation based on inverse rendering \cite{alotaibi2017biophysical}.
Our method achieves comparable performance to physics-based skin models \cite{gitlina2020practical, aliaga2022estimation},
without requiring the calculation of such complex interactions,
as shown in Fig.~\ref{fig:exp_comparison_augment}.

\section{Ablation Study}
\label{sec:ablation}

\begin{figure*}
    \centering
    \includegraphics[width=\linewidth]{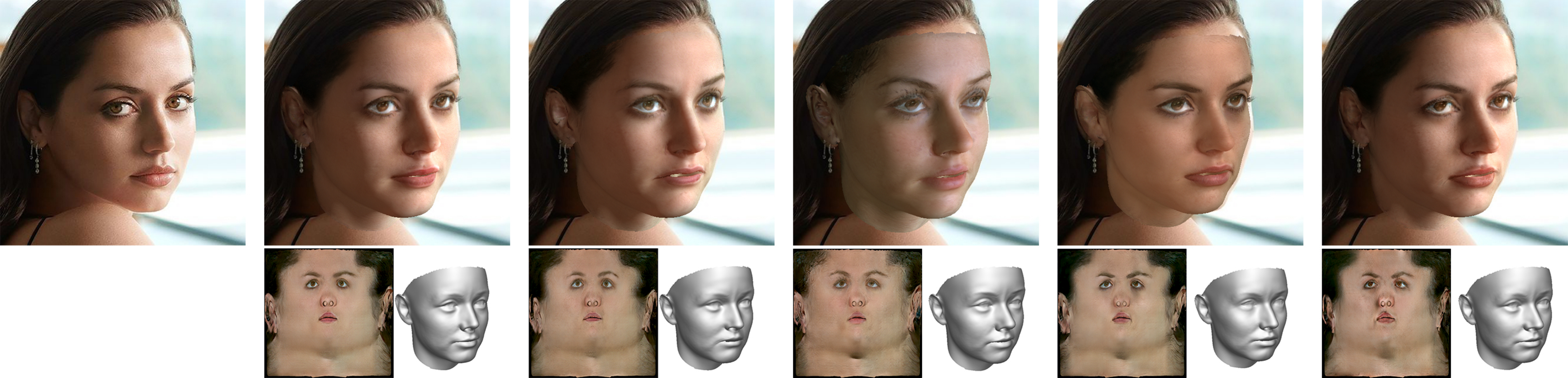}
    \makebox[0.1666\linewidth][c]{\footnotesize{a) Input}}\hfill
    \makebox[0.1666\linewidth][c]{\footnotesize{b) Ours}}\hfill
    \makebox[0.1666\linewidth][c]{\footnotesize{c) no $\mathcal{L}_{ID}$}}\hfill
    \makebox[0.1666\linewidth][c]{\footnotesize{d) no $\mathcal{L}_{per}$}}\hfill
    \makebox[0.1666\linewidth][c]{\footnotesize{e) no $\mathcal{L}_{ph}$}}\hfill
    \makebox[0.1666\linewidth][c]{\footnotesize{f) no $\mathcal{L}_{\mathbf{W}}$}}\hfill
    \\
    
    \caption{
        Ablation study on the losses of the \textbf{GAN inversion} optimization.
        For each example, we remove the shown loss from the optimization,
        and show the rendered result $\mathbf{I}_{R}$ (top), 
        the diffuse albedo $\mathbf{A}_D$ (bottom-left) and the shape $\mathbf{S}$ (bottom-right).
        It is apparent how all the proposed losses are required,
        in order to obtain facial shape and reflectance with high identity and visual similarity, while also maintaining an albedo without scene illumination.
    }
    \label{fig:ablation_loss}
\end{figure*}
\begin{figure*}
    \centering
    \includegraphics[width=\linewidth]{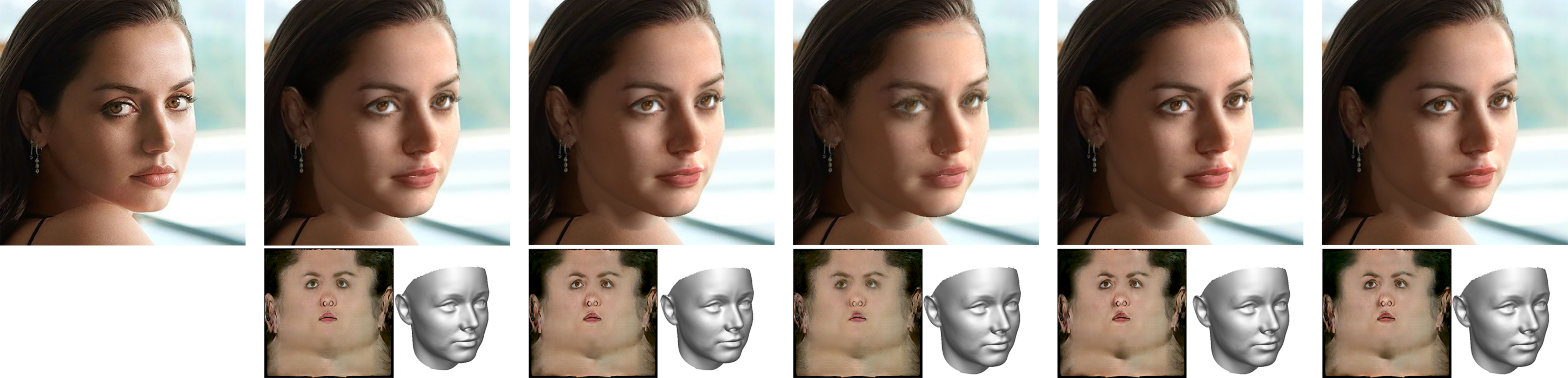}
    \makebox[0.1666\linewidth][c]{\footnotesize{a) Input}}\hfill
    \makebox[0.1666\linewidth][c]{\footnotesize{b) Ours}}\hfill
    \makebox[0.1666\linewidth][c]{\footnotesize{c) no $\mathcal{L}_{ph}$}}\hfill
    \makebox[0.1666\linewidth][c]{\footnotesize{d) no $\mathcal{L}_{LP}$}}\hfill
    \makebox[0.1666\linewidth][c]{\footnotesize{e) no $\mathcal{L}_{flip}$}}\hfill
    \makebox[0.1666\linewidth][c]{\footnotesize{f) no $\mathcal{L}_{\kappa}$}}\hfill
    \\
    
    \caption{
        Ablation study on the losses of the \textbf{GAN tuning} optimization.
        For each example, we remove the shown loss from the optimization,
        and show the rendered result $\mathbf{I}_{R}$ (top), 
        the diffuse albedo $\mathbf{A}_D$ (bottom-left) and the shape $\mathbf{S}$ (bottom-right).
        Despite most cases showing a highly optimized rendered result,
        without the proposed losses combination,
        the albedo absorbs residual scene illumination, not captured by our rendering,
        in cases c), e), f),
        Also, since we do not optimize the shape $\mathbf{S}$ during tuning,
        the shape remains the same in these examples.
    }
    \label{fig:ablation_loss_pti}
\end{figure*}
\begin{figure*}
    \centering
    \includegraphics[width=\linewidth]{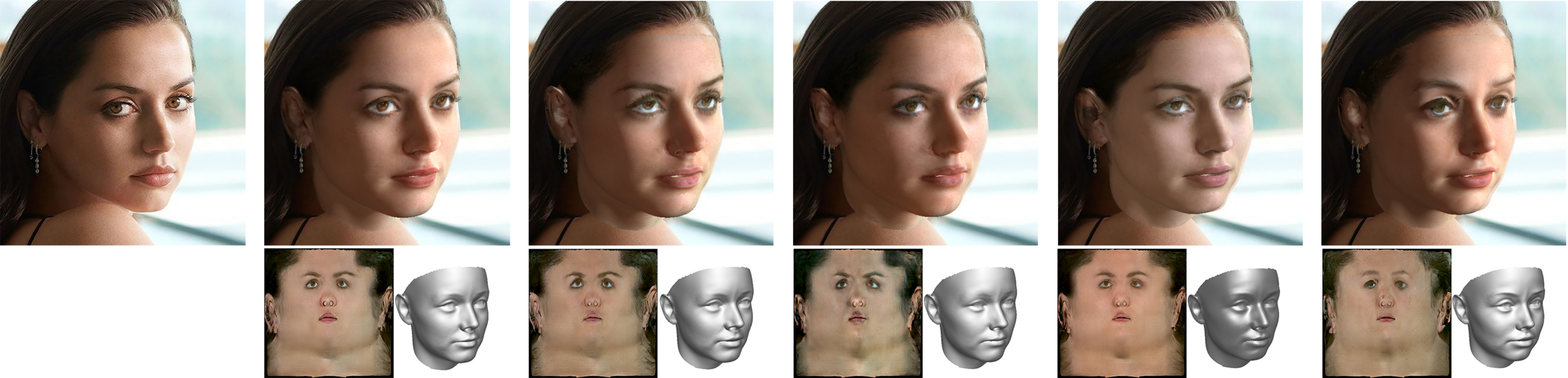}
    \makebox[0.1666\linewidth][c]{\footnotesize{a) Input}}\hfill
    \makebox[0.1666\linewidth][c]{\footnotesize{b) Ours}}\hfill
    \makebox[0.1666\linewidth][c]{\footnotesize{c) no GAN tuning}}\hfill
    \makebox[0.1666\linewidth][c]{\footnotesize{d) $\mathbf{W}+$ instead of $\mathbf{W}$ \cite{luo2021normalized}}}\hfill
    \makebox[0.1666\linewidth][c]{\footnotesize{e) Pytorch3D \cite{ravi2020pytorch3d} }}\hfill
    \makebox[0.1666\linewidth][c]{\footnotesize{f) $\mathbf{z}$ fitting instead of $\mathbf{W}$}}\hfill
    \\
    
    \caption{
        Ablation study on \textbf{method choices}.
        For each example, make the shown change during the optimization,
        and show the rendered result $\mathbf{I}_{R}$ (top), 
        the diffuse albedo $\mathbf{A}_D$ (bottom-left) and the shape $\mathbf{S}$ (bottom-right).
        c), e)and f) show the importance of $\mathbf{W}$ fitting,
        our photorealistic facial shading and the GAN tuning.
        d) shows that $\mathbf{W+}$ optimization, as performed by \cite{luo2021normalized}, achieves great likeness in the albedo,
        but does not maintain the semantics of the facial reflectance.
    }
    \label{fig:ablation_method}
\end{figure*}

We perform three ablation studies to validate our architectural and optimization choices.
The first ablation, shown in Fig.~\ref{fig:ablation_loss},
shows the effect of the losses used in the GAN inversion part of our method.
For each example, we remove one of the following losses:
the identity loss $\mathcal{L}_{ID}$, the perceptual loss $\mathcal{L}_{per}$,
the photometric loss $\mathcal{L}_{ph}$ and the $\mathbf{W}$ regularization loss $\mathcal{L}_{\mathbf{W}}$.
Removing the landmark loss $\mathcal{L}_{lan}$ or
the shape and expression regularization losses $\mathcal{L}_{s}, \mathcal{L}_{e}$,
fails the optimization, as the shape fitting is misplaced.

Moreover, in Fig.~\ref{fig:ablation_loss_pti},
we perform a similar ablation for the losses used during the GAN tuning part of our method.
The removal of the photometric $\mathcal{L}_{ph}$ and LPIPS $\mathcal{L}_{LP}$ losses,
shows their importance in achieving higher likeness to the input image,
The removal of the flip $\mathcal{L}_{flip}$ and chromaticity $\mathcal{L}_{\kappa}$ losses
shows their need in maintaining an albedo without scene illumination,
when compared to pivotal tuning \cite{roich2022pivotal}.

Finally, in Fig.~\ref{fig:ablation_method},
we show an ablation study on choices regarding the optimization method.
In Fig.~\ref{fig:ablation_method} c),
we show the optimized result,
without our GAN tuning,
Fig.~\ref{fig:ablation_method} d)
shows an extended latent space $\mathbf{W+}$ optimization,
rather than $\mathbf{W}$, which means that each slice of $\mathbf{W}$
is separately optimized for each layer of the generator network
and is suggested by \cite{luo2021normalized}.
Such an approach also achieves great likeness in the optimized rendering,
but disturbs the statistics of the generator,
so that scene illumination is absorbed by the albedo,
and the network cannot be accurately manipulated (Fig.~\ref{fig:pca_manip}).
Similar observations are also reported by \cite{roich2022pivotal}.
In Fig.~\ref{fig:ablation_method} e)
we show the full optimization (inversion and tuning),
while using the Pytorch3D \cite{ravi2020pytorch3d} implementation of Phong shader,
which also only supports the diffuse albedo $\mathbf{A}_D$,
no spatial variation in specular roughness and no subsurface scattering optimization.
Fig.~\ref{fig:ablation_method} f)
shows the optimization of the latent variable $\mathbf{z}$ which is passed through a mapping network before given to the generator. Such a fitting is more constrained and cannot reach a satisfying identity similarity.

\section{Additional Results}
\label{sec:comparisons}

\begin{figure*}[h]
    \centering
    \includegraphics[width=0.9\linewidth]{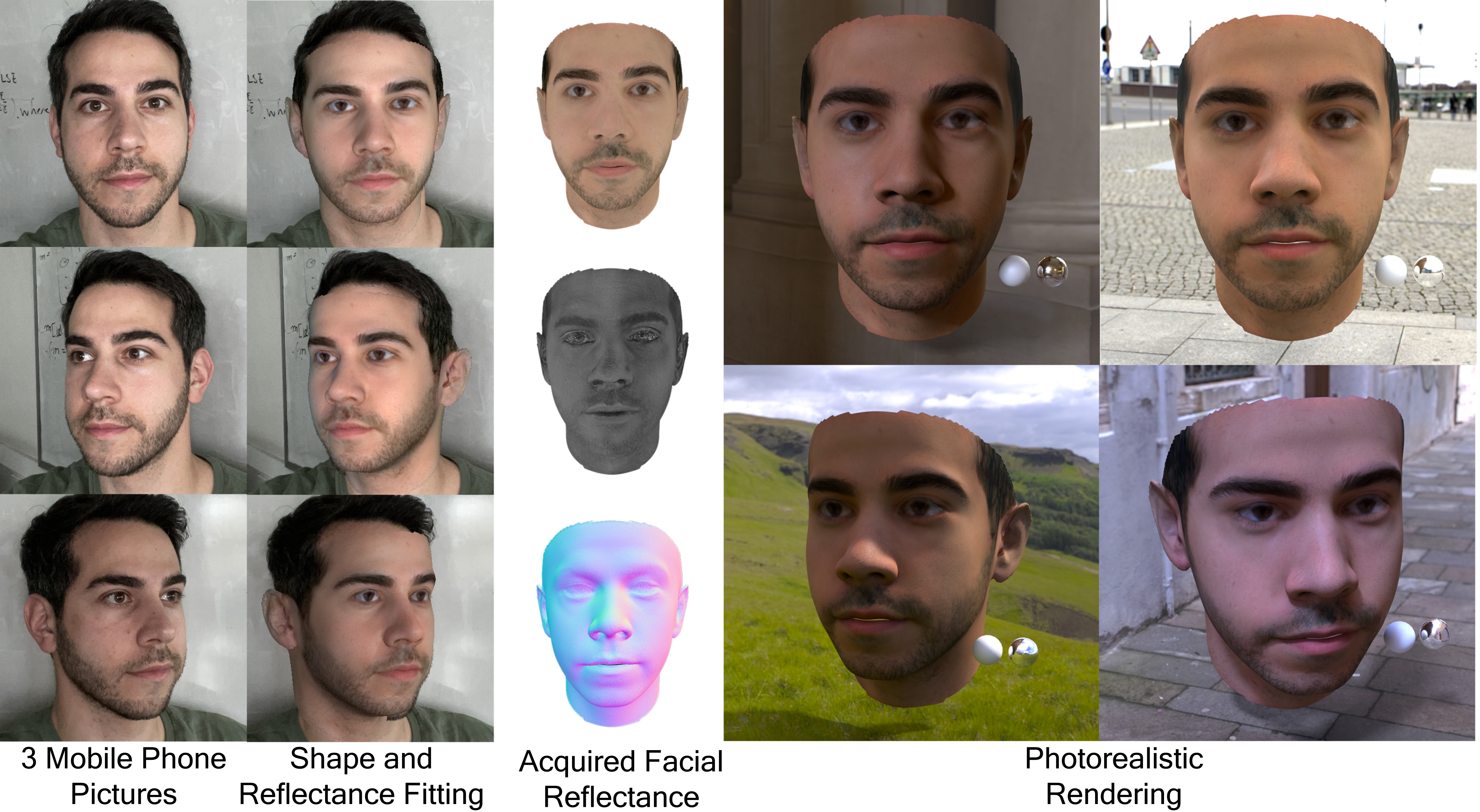}\\
    \caption{
        Additional multi-view capture results, using our method.
        From left to right:
        a) three input images,
        b) rendered fitting using our method,
        c) diffuse albedo $\mathbf{A}_D$, specular albedo $\mathbf{A}_S$ and normals $\mathbf{N}_S$,
        d) rendered shape $\mathbf{S}$
        and e) rendered results on various environments.
    }
    \label{fig:multi1}
\end{figure*}

\begin{figure*}[h]
    \centering
    \includegraphics[width=0.9\linewidth]{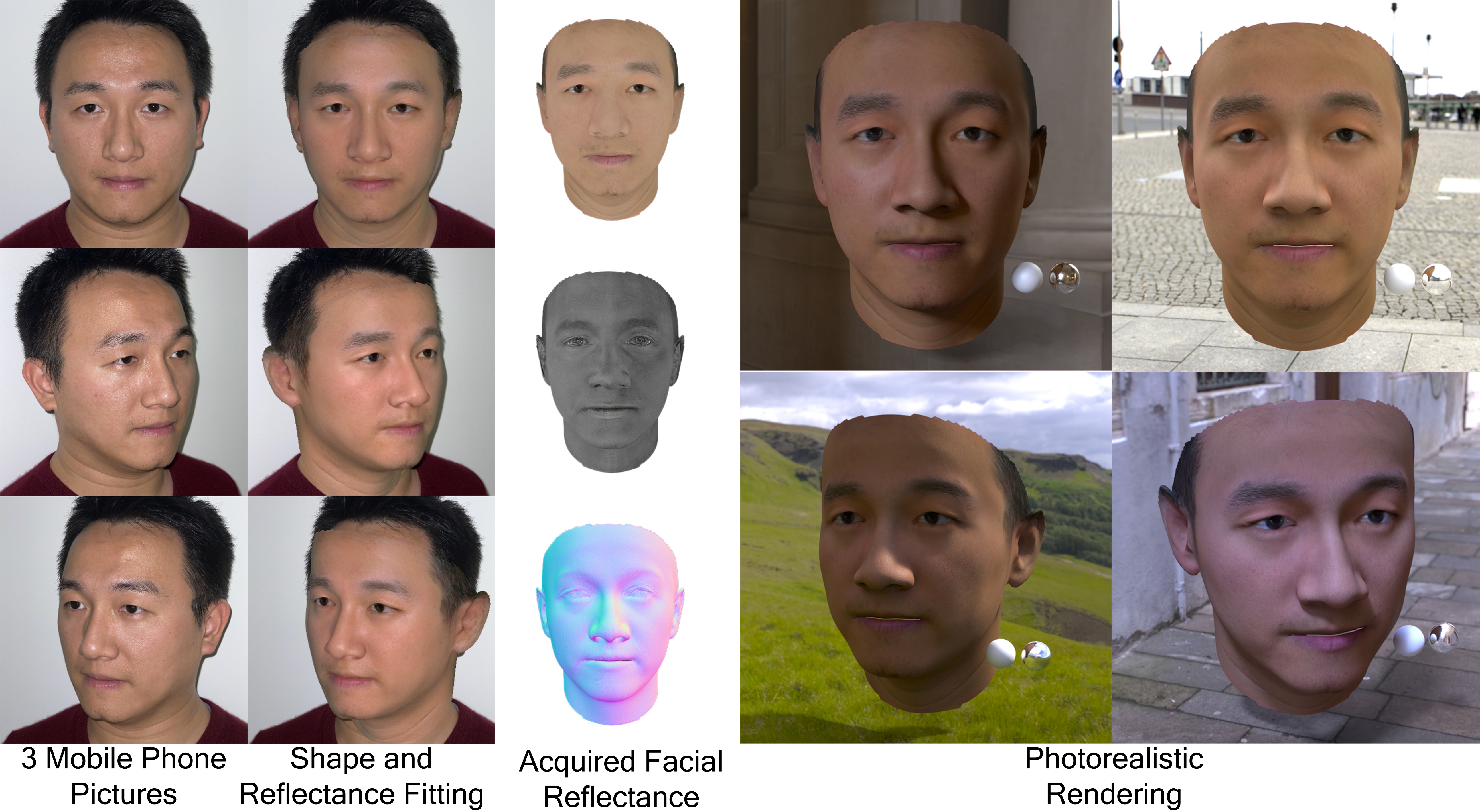}
    \caption{
        Additional multi-view capture results, using our method.
        From left to right:
        a) three input images,
        b) rendered fitting using our method,
        c) diffuse albedo $\mathbf{A}_D$, specular albedo $\mathbf{A}_S$ and normals $\mathbf{N}_S$,
        d) rendered shape $\mathbf{S}$
        and e) rendered results on various environments.
    }
    \label{fig:multi2}
\end{figure*}

\begin{figure*}[h]
    \centering
    \includegraphics[width=0.9\linewidth]{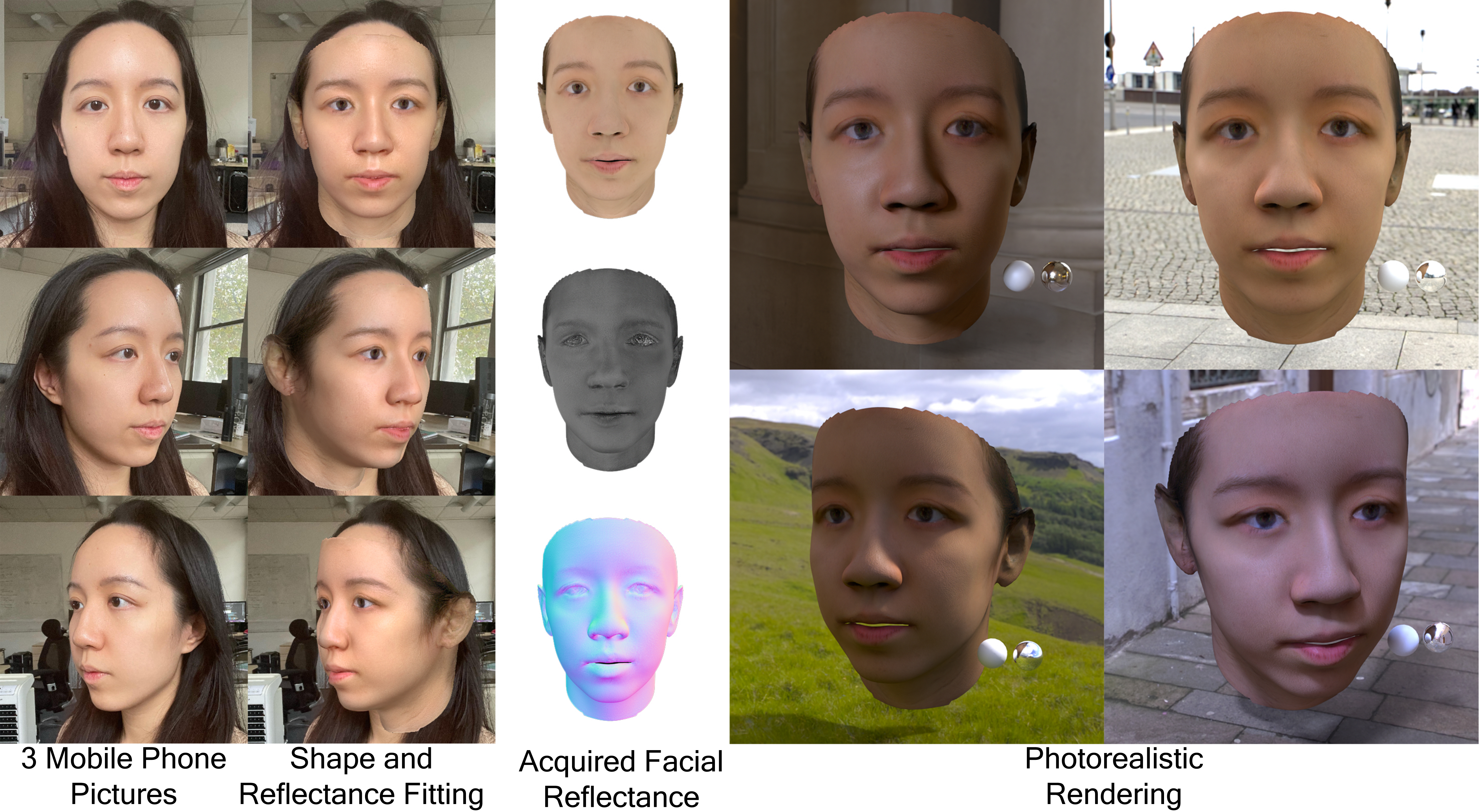}
    \caption{
        Additional multi-view capture results, using our method.
        From left to right:
        a) three input images,
        b) rendered fitting using our method,
        c) diffuse albedo $\mathbf{A}_D$, specular albedo $\mathbf{A}_S$ and normals $\mathbf{N}_S$,
        d) rendered shape $\mathbf{S}$
        and e) rendered results on various environments.
    }
    \label{fig:multi3}
\end{figure*}

We present additional results of multi-view capturing in Figs.~\ref{fig:multi1},\ref{fig:multi2},\ref{fig:multi3}.
In each case we ask the subject to take three unconstrained mobile phone images,
from the front and the side.
We show that these are enough to create an accurate photorealistic avatar of the subject,
using our method.
In total, capturing takes less than 10 seconds and processing less than a minute.

Additional to the quantitative comparison on identity similarity presented
in the main manuscript,
we also perform and present here a quantitative comparison on shape reconstruction.
Following the benchmark proposed by GANFit \cite{gecer2019ganfit},
we reconstruct the 53 subjects from the MICC Florence 3D Faces dataset (MICC) \cite{Bagdanov:2011:FHF:2072572.2072597} with our facial shape
and measure their distance (in nm) from the dataset shapes,
using the open-source benchmark code of GANFit \cite{gecer2019ganfit}.
MICC includes three subcategories, ``cooperative'', ``indoor'' and ``outdoor'',
on which we report of findings separately.
Instead of 5 images \cite{gecer2019ganfit}, 
we only use 3 random images from each video for the reconstruction.
We then perform dense alignment and measure the point-to-plane distance
and present our findings in Tab.~\ref{tab:micc}.

Our method performs similarly in shape reconstruction to GANFit \cite{gecer2019ganfit} and GANFit++ \cite{gecer2021fast},
and significantly better when compared to previous methods \cite{genova2018unsupervised, tran2019towards, booth20163d} and Fast-GANFit \cite{gecer2021fast}.
GANFit reconstructions are slightly closer to the ground truth shapes.
This can be explained by the fact that we optimize both the shape mesh and the texture normals, 
and thus a part of the shape information is explained in the normals domain rather than in the actual shape space.
For a fair comparison, we only compared the meshes, potentially missing details.
Finally, as shown in the main manuscript, our method scores first in identity similarity,
while also acquiring relightable reflectance textures.

\begin{table*}[h]
    \centering
    \begin{tabular}{l|c|c|c}
                            &  \textbf{Cooperative}  & \textbf{Indoor}  & \textbf{Outdoor}  \\
        \textbf{Method}     &  Mean $\pm$ Std.  & Mean $\pm$ Std.   & Mean $\pm$ Std.   \\
        \hline
        Tran et al. \cite{tuan2017regressing}
                            & $1.93 \pm 0.27$   & $2.02 \pm 0.25$   & $1.86 \pm 0.24$   \\
        Booth et al. \cite{booth20163d}
                            & $1.82 \pm 0.29$   & $1.85 \pm 0.22$   & $1.63 \pm 0.16$   \\
        Genova et al. \cite{Genova_2018_CVPR}
                            & $1.50 \pm 0.14$   & $1.50 \pm 0.11$   & $1.48 \pm 0.11$   \\
        GANFit \cite{gecer2019ganfit}
                            & $0.95 \pm 0.10$   & $0.94 \pm 0.10$   & $0.94 \pm 0.10$  \\
        GANFit++ \cite{gecer2021fast}
                            & $0.94 \pm 0.17$   & $0.92 \pm 0.14$   & $0.94 \pm 0.19$   \\
        Fast-GANFit \cite{gecer2021fast}
                            & $1.11 \pm 0.25$   & $0.98 \pm 0.15$   & $1.16 \pm 0.18$   \\
        Ours                & $0.95 \pm 0.18$   & $0.97 \pm 0.20$   & $0.98 \pm 0.21$    
    \end{tabular}
    \caption{
        Quantitative benchmark comparison of shape reconstruction,
        on the MICC Florence 3D Faces dataset,
        using point-to-plane distance,
        as used in the open-source benchmark of \cite{gecer2021fast}.
        we compare our results with the reported results of
        Tran et al.~\cite{tuan2017regressing},
        Booth et al.~\cite{booth20163d},
        Genova et al.~\cite{Genova_2018_CVPR},
        GANFit \cite{gecer2019ganfit},
        GANFit++ \cite{gecer2021fast} and Fast-GANFit \cite{gecer2021fast}.
    }
    \label{tab:micc}
\end{table*}

\section{Implementation Details}
\label{sec:details}

As briefly described in the main manuscript,
\ModelName{} implementation builds on the public repository of StyleGAN2-ADA \cite{karras2020training},
in pytorch, both for the generator and the discriminator.
However, we make the following changes.
a) The generator $\mathcal{G}$ is branched on the last convolutional blocks,
which is achieved feeding the output of the last single-branch layer,
to 3 different copies of the last branched module.
The generator follows the skip-connections architecture of \cite{karras2020training}.
b) The discriminator is also branched, and follows the resnet architecture of \cite{karras2020training}.
The output of each branch is concatenated, and fed to the last convolutional block and the fully connected layers.

For the differentiable photorealistic rendering, we create a shader based on the Blinn-Phong model \cite{blinn1977models}, following AvatarMe++\cite{lattas2021avatarme++}.
The implementation of the model is done by extending the shader and shading classes in Pytorch3D \cite{ravi2020pytorch3d}.

Our optimization is based on a number of hyperparameters $\lambda_i$,
each corresponding to a loss described in the paper.
We find these empirically, and present them below.
For the inversion, we use a learning rate of
$l_{inv}=1\times{}10^{-2}$ and for the tuning we use a learning rate of
$l_{pti}=8\times{}10^{-4}$.
In Table~\ref{tab:lambda_inversion} and Table~\ref{tab:lambda_tuning}
we present the values which we find most optimal,
when optimizing for crop and rendering at $512\times512$ pixels,
250 iterations for inversion and 30 iterations for tuning.

\begin{table*}[h]
    \centering
    \begin{tabular}{r|l|l|l|l|l|l|l}
        \textbf{hyper-param} &
        $\lambda_1$         & $\lambda_2$           & $\lambda_3$           & $\lambda_4$           &
        $\lambda_5$         & $\lambda_6$           & $\lambda_7$           \\
        \hline
        \textbf{corresponding loss} &
        $\mathcal{L}_{lan}$  & $\mathcal{L}_{ph}$   & $\mathcal{L}_{ID}$    & $\mathcal{L}_{per}$    &
        $\mathcal{L}_{W}$    & $\mathcal{L}_{s}$    & $\mathcal{L}_{e}$     \\
        \hline
        \textbf{value} &
        $100$               & $0.5$                 & $1.0$                   & $25.0$                  &
        $5\times{}10^{-2}$           & $5\times{}10^{-4}$             & $5\times{}10^{-4}$             \\   
    \end{tabular}
    \caption{
        Hyper-parameter values and their corresponding loss terms,
        used for the GAN-inversion part of the proposed method.
    }
    \label{tab:lambda_inversion}
    \centering
    \begin{tabular}{r|l|l|l|l}
        \textbf{hyper-param} &
        $\lambda_8$         & $\lambda_9$           & $\lambda_{10}$           & $\lambda_{11}$           \\
        \hline
        \textbf{corresponding loss} &
        $\mathcal{L}_{LP}$  & $\mathcal{L}_{ph}$    & $\mathcal{L}_{flip}$  & $\mathcal{L}_{\kappa}$ \\
        \hline
        \textbf{value} &
        $2.0$                 & $0.5$                 & $0.8$                 & $0.35$
    \end{tabular}
    \caption{
        Hyper-parameter values and their corresponding loss terms,
        used for the GAN-tuning part of the proposed method.
    }
    \label{tab:lambda_tuning}
\end{table*}

\newpage

\end{document}